\pdfoutput=1
\documentclass[11pt,a4paper]{article}
\usepackage[table]{xcolor}
\usepackage[acceptedwithA]{tacl2021v1}
\usepackage{times}
\usepackage{latexsym}

\usepackage[T1]{fontenc}
\usepackage[utf8]{inputenc}

\usepackage{microtype}

\usepackage{inconsolata}

\usepackage{graphicx}
\usepackage{array}

\usepackage{CJKutf8}

\usepackage{hyperref}
\usepackage{url}

\usepackage{amssymb}
\usepackage{amsmath}
\usepackage{amsthm}
\usepackage{amsfonts}
\usepackage{bm}
\usepackage{nicefrac}
\usepackage{dsfont}

\usepackage{booktabs}
\usepackage{multirow}
\usepackage{makecell}
\usepackage{arydshln}

\usepackage{color}
\usepackage{colortbl}

\usepackage{comment}
\usepackage{xspace}
\usepackage{soul}
\usepackage{subcaption}

\usepackage{enumitem}
\usepackage[normalem]{ulem}
\setlist[itemize,enumerate]{leftmargin=*}

\definecolor{light-orange}{HTML}{fee9d4}
\definecolor{light-green}{HTML}{d8f0d3}
\definecolor{light-blue}{HTML}{dae8f5}
\definecolor{set10-red}{HTML}{e41a1c}
\definecolor{set10-blue}{HTML}{377eb8}
\definecolor{set10-green}{HTML}{4daf4a}

\usepackage{CJKutf8}

\usepackage{hyperref}
\usepackage{url}

\usepackage{amssymb}
\usepackage{amsmath}
\usepackage{amsthm}
\usepackage{amsfonts}
\usepackage{bm}
\usepackage{nicefrac}
\usepackage{dsfont}
\usepackage{fancyvrb}
\usepackage{fvextra}
\usepackage{listings}
\usepackage{xcolor}
\usepackage{tabularx}% added for table design
\usepackage[all]{nowidow}
\usepackage{xcolor}
\usepackage{colortbl}

\usepackage{booktabs}
\usepackage{multirow}
\usepackage{makecell}
\usepackage{arydshln}

\usepackage{comment}
\usepackage{xspace}
\usepackage{soul}

\usepackage{enumitem}
\usepackage[normalem]{ulem}
\setlist[itemize,enumerate]{leftmargin=*}

\makeatletter
\def\adl@drawiv#1#2#3{%
        \hskip.5\tabcolsep
        \xleaders#3{#2.5\@tempdimb #1{1}#2.5\@tempdimb}%
                #2\z@ plus1fil minus1fil\relax
        \hskip.5\tabcolsep}
\newcommand{\cdashlinelr}[1]{%
  \noalign{\vskip 2pt
           \global\let\@dashdrawstore\adl@draw
           \global\let\adl@draw\adl@drawiv}
  \cdashline{#1}[.4pt/2pt]
  \noalign{\global\let\adl@draw\@dashdrawstore
           \vskip 2pt}}
\makeatother

\definecolor{light-orange}{HTML}{fee9d4}
\definecolor{light-green}{HTML}{d8f0d3}
\definecolor{light-blue}{HTML}{dae8f5}
\definecolor{set10-red}{HTML}{e41a1c}
\definecolor{set10-blue}{HTML}{377eb8}
\definecolor{set10-green}{HTML}{4daf4a}

\definecolor{CustomBlue}{RGB}{57,83,191}
\definecolor{CustomRed}{HTML}{a75151}
\definecolor{DarkGreenOne}{RGB}{106,168,79}
\usepackage[most]{tcolorbox}

\newtcbox{\clustertab}[1]{on line, box align=base, colback={#1},colframe={#1},size=fbox,arc=2pt,top=-1.5pt, bottom=-1.5pt, left=-1.5pt, right=-1.5pt, boxrule=0pt, enlarge left by=1pt}

\newcommand{\mint}{\textsc{Mint}}

\newcommand{\Towervtwo}{\textsc{Tower-v2}}

\newcommand{\comet}{\textsc{Comet}}
\newcommand{\bleurt}{\textsc{Bleurt}}
\newcommand{\bleu}{\textsc{Bleu}}

\newcommand{\chrf}{\textsc{chrF}}
\newcommand{\cometkiwi}{\textsc{CometKiwi}}
\newcommand{\xcomet}{\textsc{xComet}}

\newcommand{\gptfour}{\textsc{GPT-4}}
\newcommand{\claudethreefive}{\textsc{Claude-Sonnet-3.5}}

\newcommand{\wmttwofour}{\textsc{WMT24}}
\newcommand{\wmttwothree}{\textsc{WMT23}}
\newcommand{\metricx}{\textsc{MetricX}}

\newcommand{\iscore}{\textsc{InstructScore}}
\newcommand{\prometheus}{\textsc{Prometheus 2}}
\newcommand{\gemma}{\textsc{Gemma}}
\newcommand{\methodname}{\textsc{MintAdjust}}
\newcommand{\autorank}{\textsc{AutoRank}}
\newcommand{\gemini}{\textsc{Gemini-1.5-Pro}}
\newcommand{\metametric}{\textsc{MetaMetrics-MT}}

\newcommand{\andre}[1]{{\textcolor{red}{[AM: #1]}}}

\title{Adding Chocolate to \mint{}\mintleaf{}: \\ Mitigating Metric Interference in Machine Translation}
\author{
  José Pombal$^{1,2,3}$, Nuno M. Guerreiro$^{1,2,3,4}$, Ricardo Rei$^{1}$, André F. T. Martins$^{1,2,3,5}$
  \\
  \ \\
  $^1$Unbabel, $^2$Instituto de Telecomunicações
  \\
  $^3$Instituto Superior Técnico, Universidade de Lisboa 
  \\
  $^4$MICS, CentraleSupélec, Université Paris-Saclay, $^5$ELLIS Unit Lisbon
  \\
  \texttt{jose.pombal@unbabel.com}
}

\definecolor{CustomBlue}{RGB}{57,83,191}
\definecolor{CustomRed}{HTML}{B51935}
\definecolor{Custom}{HTML}{B74EB4}
\definecolor{CustomGreen}{HTML}{208f0d}

\begin{document}
\maketitle
\begin{abstract}
%Automatic metrics are often leveraged during training and inference to enhance the performance of large language models.
%
As automatic metrics become increasingly stronger and widely adopted, the risk of unintentionally ``gaming the metric'' during model development rises.
This issue is caused by metric interference (\mint{}), \textit{i.e.}, the use of the same or related metrics for both model tuning and evaluation.
\mint{} can misguide practitioners into being overoptimistic about the performance of their systems: as system outputs become a function of the interfering metric, their estimated quality loses correlation with human judgments.
In this work, we analyze two common cases of \mint{} in machine translation-related tasks: filtering of training data, and decoding with quality signals.
Importantly, we find that \mint{} strongly distorts instance-level metric scores, even when metrics are not directly optimized for---questioning the common strategy of leveraging a different, yet related metric for evaluation that is not used for tuning. 
To address this problem, we propose \methodname{}, a method for more reliable evaluation under \mint{}.
%
%\methodname{} takes as input scores of other metrics on outputs from models not subject to interference, and produces adjusted scores for the interfering metric.
%
On the \wmttwofour{} MT shared task test set, \methodname{} ranks translations and systems more accurately than state-of-the-art-metrics across a majority of language pairs, especially for high-quality systems.
Furthermore, \methodname{} outperforms \autorank{}, the ensembling method used by the organizers.\footnote{We will release a codebase for replicating the results in this work upon publication.}
%
%Our work contributes to improving the understanding of the effects of \mint{} on automatic evaluation and to developing better mitigation strategies.

%\pombal{I feel like I need to justify why I'm focusing on MT; do I?}
%\pombal{mention \autorank{}-Ins in the abstract?}

\end{abstract}

\section{Introduction}

\begin{figure*}
    \centering
    \includegraphics[width=0.85\linewidth]{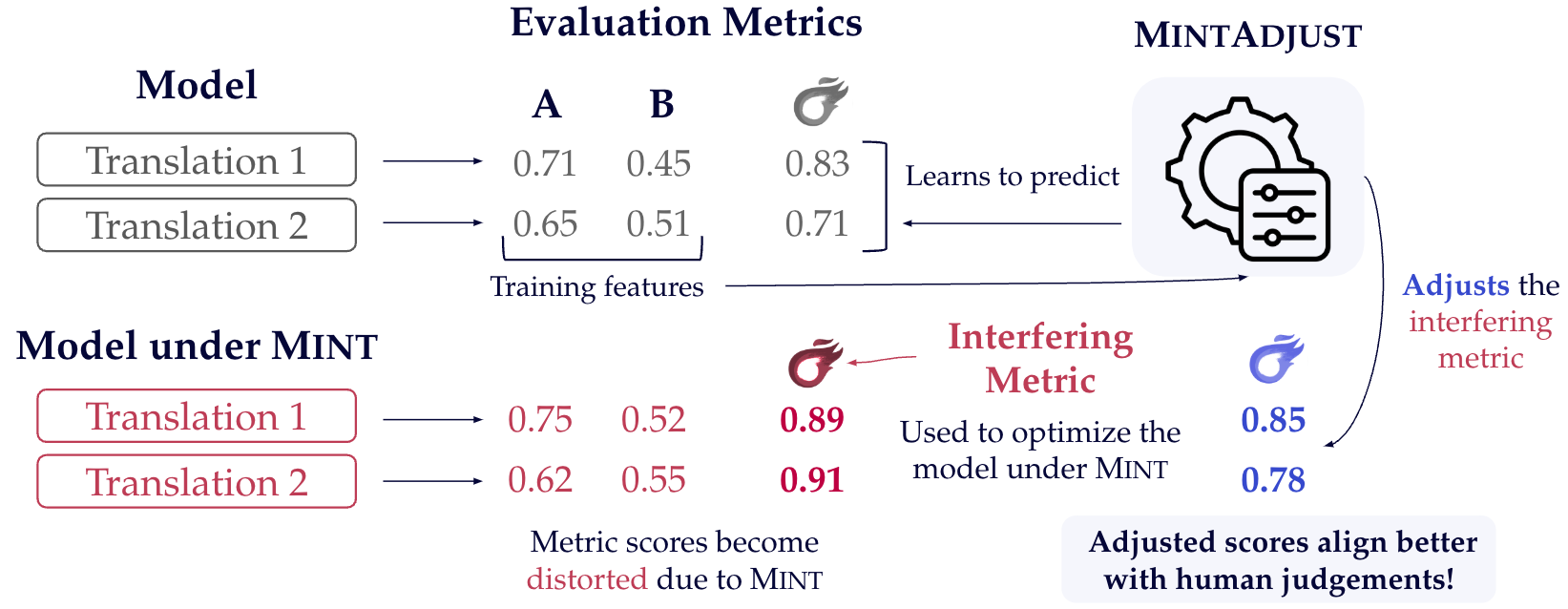}
    \caption{
    \methodname{} learns to predict the instance-level scores of an interfering metric, given the scores of other metrics measured in translations produced by models that are not under \mint{} (in \textcolor{gray}{gray}). 
    At test time, for each instance from the model under \mint{}, \methodname{} receives the \textcolor{CustomRed}{scores of other metrics}, and produces a \textcolor{CustomBlue}{new interfering metric score} that correlates better with human judgments.
    }
    \label{fig:method-fig}
\end{figure*}

Automatic evaluation metrics are evolving beyond their traditional role in model selection, now serving as tools for enhancing the performance of large language models (LLMs) across their entire development pipeline.
Applications include filtering high-quality training data, or serving as proxies for human preferences in alignment and decoding~\citep{xucontrastive,zhu2024preference,wu2024better}.
In the field of machine translation (MT), several works have leveraged such methods to claim state-of-the-art results~\citep{fernandes2022quality,xucontrastive,rei2024tower}.
In these cases, the same or similar metrics to the one used for optimization are also used downstream for evaluation---we call this phenomenon \textbf{metric interference} (\mint{}).

An unintended consequence of \mint{} is that it may jeopardize evaluation.
For example, an interfering metric can lose much of its correlation with human judgements when ranking systems~\cite{fernandes2022quality}, or make the optimized model generate absurd outputs that the metric scores highly~\citep{yan2023bleurt}.
At worst, authors disregard \mint{} and use the same metric for optimization and evaluation.
At best, as recommended by~\citet{kocmi-etal-2024-navigating}, they acknowledge it and perform evaluation with another metric~\citep{fernandes2022quality,rei2024tower,kocmi2024findings}.
The latter solution, however, may not work well if metrics correlate strongly among each other, as they can have similar biases.
%
%In MT, there is reason to believe this is the case because metrics often have similar architectures and are trained on the same data~\citep{kovacsmitigating}.
%
%A notable recent example of problematic \mint{} is \Towervtwo{}, the winning submission of the \wmttwofour{} MT Shared Task~\citep{kocmi2024findings}.
%
%The system used minimum Bayes risk decoding~\citep[MBR]{freitag2022high} with \comet{}~\citep{rei2020comet} and beat \claudethreefive{}, the best system overall, on all language pairs (LPs) according to another automatic metric, \metricx{}~\citep{juraska-etal-2023-metricx}, but beat it only on 5 of 11 LPs after human evaluation.
%
%Lexical metrics are the exception, as they are known to correlate negatively at a system level with neural metrics, when optimizing for the latter \tocite{QAD}.
%
%However, the driver of this negative correlation, and other implications of \mint{} for model selection and metric score interpretation are not well understood.
% In general, the effects of \mint{} on model evaluation and the effectiveness of the aforementioned mitigation techniques are not well understood.
%
The effects of \mint{} are not well understood, and existing mitigation methods lack effectiveness.

In the first part of our work, we investigate the implications of \mint{} for evaluation with a series of metrics on two common \mint{} cases in MT: 1) data filtering, where a metric is used to select high-quality data from a larger pool to more efficiently train a translation model; 2) MBR, a widely-used decoding strategy that uses a metric to select the best translation from a pool of candidates.
%These correlations hold up in common cases of \mint{}: training data filtering with quality estimation metrics, and MBR decoding with reference-based metrics.
%
%However, while most metrics agree on the best model under data filtering, that is not the case with MBR decoding, a presumably worse case of \mint{}.
%
%Comparing greedy decoding and MBR, we find that neural and lexical metrics disagree when ranking systems, but agree on the direction of change in quality of the majority of translations (instances).
%
%The system-level negative correlation is driven by a relative minority of instances where metrics disagree.
%
We find that the effects of \mint{} are stronger for MBR than for data filtering.
Specifically, the instance-level scores of metrics are heavily distorted, to the point where large improvements in an interfering metric become consistently related to deterioration in other metrics.
Importantly, similar distortions hold when comparing an interfering metric with human scores, explaining the system-level findings of~\citep{fernandes2022quality}.
Furthermore, the scores of metrics that correlate with the interfering metric will also be overestimated.
%---this is the case for most state-of-the-art MT metrics.
%
%Such effects are also present, though less pronounced, in the data filtering setting.
%
This brings into question the common practice of using a single related metric for evaluation.

Building on our findings, we propose a method for correcting an interfering metric's scores by learning its distribution without \mint{} from a large set of other metrics, as opposed to relying on a single one: \methodname{} (summarized in Figure~\ref{fig:method-fig}).
Using WMT24 human-judged translations, we show that, under \mint{}, \methodname{} ranks systems and individual translations more accurately than the interfering metric and other state-of-the-art metrics on a majority of language pairs.
%
%Furthermore, it consistently ranks individual translations more accurately than alternatives.
%
Its performance is particularly strong when comparing high-quality MT systems.
Our correction method generalizes across translation models and metrics---\textit{i.e.}, it can be learned from a given set of models and metrics and applied to a different model. 
%-- posing a better alternative for model selection than other metrics in isolation.
%
%Furthermore, motivated by the importance of individual instances in \mint{}, we propose \autorank{}-Ins, an instance level variant of \autorank{}, the ensembling method used by~\citet{kocmi2024findings} to rank systems with automatic metrics.
%
%\autorank{}-Ins outperforms \autorank{} at a system-level and performs similarly to \methodname{} at an instance-level.

Through our fine-grained analysis and the proposed methods, we lay the groundwork for better evaluation practices in settings where automatic metrics are used for both evaluation and model optimization. 
Our work has implications not only for MT but for evaluation in general.

\section{Metric Interference}\label{sec:metric-interference}

When used for evaluation, most automatic metrics $\mathcal{M}$, may be described as functions that produce a judgement $\hat{q}$ given an input $x$, a model's output $h$, and a reference output $r$:\footnote{In translation, some metrics only use $x$ and $h$ (\textit{e.g.}, quality estimation), or $h$ and $r$ (\textit{e.g.}, lexical metrics)}
\begin{equation}
    \hat{q} = \mathcal{M}(x, h(x), r).
\end{equation}
%
%In this paper, we focus on real-valued judgements and textual inputs, model outputs, and references, but this need not be the case.

\mint{} occurs when $\mathcal{M}$---the \textbf{interfering metric}---influences a model's output, directly or indirectly, through its judgements. Thus, during evaluation, $\mathcal{M}$ becomes a function of itself:

\begin{equation}
    \hat{q} = \mathcal{M}(x, h(x; \mathcal{M'}), r)
\end{equation}
$\mathcal{M'}$ and $\mathcal{M}$ can be the same or spuriously correlated metrics---we will show later that \mint{} also affects the latter.
\textbf{Direct interference} occurs when a metric is used to choose among model outputs at test time; for example, in MBR the best candidate out of a pool of model generations is chosen by a metric.
\textbf{Indirect interference} happens when the metric influences the model's outputs by influencing the model's weights, $\theta$, through either data or tuning: $\hat{q} = \mathcal{M}(x, h(x; \theta(\mathcal{M})), r)$. 
Training data filtering is a case of indirect interference, whereby the metric influences the model's weights by changing its training distribution.
Preference optimization is an example of indirect interference through tuning.
We posit the negative effects of \mint{} on evaluation are stronger for direct interference.
%
%Figure~\ref{fig:mint-taxonomy}\pombal{WIP} summarizes this taxonomy.

%We posit that the negative effects of direct interference on evaluation are stronger than indirect interference.
%
%We verify this claim to a certain extent, but leave further exploration for future work.
%
%We do not study preference optimization with metrics, for instance, which constitutes indirect interference but can have strong adverse effects on evaluation~\citep{gisserot2024preference}.

To the best of our knowledge, \mint{} has never been formally defined but its effects have been studied empirically to some extent.
For example, LLM judges tend to favor outputs from their own model family~\citep{panickssery2024llm,verga2024replacing}.
In MT, this is also the case with LLMs~\citep{agrawal-etal-2024-automatic-metrics} and more traditional neural metrics: models optimized for \bleurt{} can produce absurd translations that score highly~\citep{yan2023bleurt}, while metric-based decoding strategies (e.g., MBR) reduce metric correlation with human judgments at the system level~\citep{fernandes2022quality}.

A typical solution to \mint{} is using a different metric, $\mathcal{M}'$, for evaluation.
The validity of this approach hinges on the often omitted assumption that $\mathcal{M}$ and $\mathcal{M}'$ do not have spurious correlations beyond their correlations with human judgments.
We posit that this does not hold in MT, since most metrics are trained on similar data and architectures and so may share similar behaviours, which can be beneficial or detrimental for evaluation.
%\pombal{add andré's causal graph somewhere here?}

%In Section~\ref{sec:analysis}, we empirically observe that: 1) the consequences of \mint{} are more noticeable for direct interferences (e.g., MBR); 2) the effects of \mint{} are visible in evaluation metrics that correlate with the interfering metric, bringing into question common practices to mitigate \mint{}.
%
%We analyze the instance-level effects that drive the system-level phenomena reported by \citet{fernandes2022quality}, contributing towards a finer-grained understanding of \mint{}.
%
%The latter motivates our proposal of a metric that has better system-/instance-level accuracy than using the original metric or alternative metrics in isolation.

\section{Experimental Setting}\label{sec:setting}

%In this section we outline the \mint{} settings we will analyze and discuss in Section~\ref{sec:analysis}.
%
%Our general goal is to understand the effects of \mint{} on evaluation at system and instance levels, considering various interference and evaluation metrics.
%
%We focus on an indirect setting~--~training data filtering~-- and a direct one~--~MBR.
%
%Section~\ref{sec:adjusted-metric} has a different experimental setting that is detailed there.

\subsection{Overview}\label{subsec:eval} 
We analyze four language pairs (LPs) of two scripts where high-quality test data is available: English$\leftrightarrow$German and English$\leftrightarrow$Chinese.
We use the \wmttwothree{} test set and a diverse suite of reference-based, reference-free, neural, and lexical metrics for evaluation: \comet{}-22~\citep{rei-etal-2022-comet}, \cometkiwi{}~\citep{rei-etal-2022-cometkiwi}, \xcomet{}-XL-REF and QE~\citep{guerreiro2023xcomet}, \metricx{}-XL-REF and QE~\citep{juraska-etal-2023-metricx}, \bleurt{}~\citep{sellam2020bleurt}, \chrf{}~\citep{popovic2015chrf}, and \bleu{}~\citep{papineni2002bleu}.
We also present additional findings using LLM Judges~\prometheus{} 7B~\citep{prometheus} and \iscore{}~\citep{instructscore}.
%for evaluation and for MBR.

\subsection{Training Data Filtering} 
Quality Estimation~\citep[QE]{specia2010machine} is the task of assessing the quality of a translation without a reference.
Using QE metrics to filter high-quality training data from a larger pool---as opposed to training models on the entire pool or a random sample---is an efficient method for building stronger models for translation~\citep{peter-etal-2023-theres,alves2024tower}.
However, using a metric for data filtering might lead to bias in evaluation.
For example, a model trained on \cometkiwi{}-filtered data may produce translations that \cometkiwi{} and \comet{} score higher than those of a model trained on \metricx{}-filtered data only because it has learned properties that the former metrics prefer, but that the latter metric and humans do not.

To investigate whether this is the case, we fine-tune \gemma{}-2B~\citep{gemma} models\footnote{We use \gemma{}-2B because it is a small (practical for finetuning), strong LLM that is not tailored for translation, so the effects of data filtering are more likely to be visible.} on 80,000 examples from a pool of 10 million examples\footnote{80k instances is a typical dataset size for fine-tuning LLMs for translation~\citep{xuparadigm,alves2024tower}, and there are diminishing marginal returns in performance to increasing it. Sampling from a much larger pool is realistic and it allows us to measure the effects of filtering with different metrics more easily (the larger the pool, the more likely metrics are to select different instances).} sourced from OPUS~\citep{tiedemann2012parallel}, sampled in 4 different ways: 1) random sampling (baseline); 2) top-k \cometkiwi{} (CKiwi) filtering; 3) top-k \metricx{}-XL-QE (MX) filtering; 4) top-k \xcomet{}-XL-QE (xC) filtering.\footnote{We provide full details on the metrics and checkpoints used in Appendix Section~\ref{apx:metric-details}.}
Table~\ref{tab:filters-overlap} reveals minimal overlap in training sets, indicating that different metrics are likely selecting translations based on distinct preferred aspects.
%
%Interestingly, while the overlap between \cometkiwi{} and \xcomet{} is almost eight times higher than random, it is much lower between \xcomet{} and \metricx{}.
%
%Translation models may learn these metric-specific features, increasing the likelihood of undesirable interference effects downstream.
\subsection{Minimum Bayes Risk Decoding}

Decoding strategies informed by quality metrics such as minimum Bayes risk (MBR) decoding consistently outperform alternatives like greedy decoding or nucleus sampling, according to humans~\citep{fernandes-etal-2022-quality, freitag-etal-2022-high, nowakowski-etal-2022-adam, farinhas-etal-2023-empirical}.
\citet{fernandes2022quality} show the effectiveness of MBR for LLM-based translation but warn against using the same metric for evaluation due to \mint{}.
%
%In particular, they show that \comet{} ranks its own MBR model higher than humans would.
%
%In this work, we expand this kind of analysis to more metrics and to the instance-level dynamics that drive system-level phenomena.
%
%Furthermore, we explore methods for correcting the metric used for MBR at evaluation.
%
We aim to expand this kind of analysis to more metrics and to the instance-level to provide more extensive insight.
Our experiments leverage \Towervtwo{}-7B, a state-of-the-art LLM for machine translation.
We generate pools of 20 candidates through $\epsilon$-sampling~\cite{freitag-etal-2023-epsilon} with $\epsilon = 0.02$, and perform MBR with four metrics: \comet{}-22, \bleurt{}, \chrf{}, and \bleu{}.
Greedy decoding serves as the baseline.
We replicate the analysis with 50 candidates and \Towervtwo{}-70B to assess the impact of model and pool sizes and reach similar conclusions (see Tables~\ref{tab:apx-mbr-50-cands} and~\ref{tab:apx-mbr-20-70b}).

\iffalse
Throughout Section~\ref{sec:analysis}, we will focus \andre{we focus} on two kinds of analysis.
% 
\andre{move this text to sec 4?} First, we look at system-level measurements of diverse evaluation metrics, in order to understand how \mint{} affects system rankings and, conseuqently, model selection.
%
Second, we compare instance-level deltas of metrics between models that suffer from metric interference and their respective baselines. 
%
We wish to answer questions like ``How much does a 1 point \chrf{} increase in an instance correspond to \comet{} under an unbiased setting and a setting with \mint{}?''.
%
This helps us understand the underlying dynamics in \mint{}, namely the instance-level phenomena behind system-level findings, ultimately informing a better strategy for evaluation under \mint{}.
\fi

\section{Analysis}\label{sec:analysis}

We focus on English$\rightarrow$Chinese, where the phenomena we describe are more pronounced.
Similar conclusions hold for other LPs (see Tables~\ref{tab:apx-df-en-de} to~\ref{tab:apx-mbr-zh-en}).

%\pombal{consider changing to zh-en? bleurt-comet disagreements on MBR are a bit confusing and they only happen on this LP}

\subsection{Training Data Filtering}

\paragraph{Metrics prefer systems trained on data that they selected.}

Table~\ref{tab:sys-level-filter} shows system-level evaluations for data filtering.
Systems optimized for a certain metric score higher on that metric, which is consistent with findings from past literature for MBR~\citep{fernandes2022quality}.
%
%For example, performing data filtering with \metricx{} yields higher \metricx{} scores than when filtering with \cometkiwi{}, which yields higher \cometkiwi{}.
%
Reference-based versions of QE metrics also score higher on these cases (e.g., \xcomet{}-REF is highest when filtering with \xcomet{}-QE), likely because they were trained with similar architectures and data.
%
%This is not surprising since metrics of the same family are expected to correlate, as they were trained with similar architectures and data. 
%
This is a warning against evaluating with a single metric even in cases of indirect interference.

\paragraph{There are few disagreements in metric preferences.}

While the interfering metric seems to be biased in favor of itself and metrics of the same family, most metrics agree on improvements over the baseline.
Notably, both neural and lexical metrics usually improve with data filtering---contrary to past findings on MBR~\cite{fernandes2022quality}---with \cometkiwi{} filtering leading to improvements across the board on all LPs (see Tables~\ref{tab:apx-df-en-de} to~\ref{tab:apx-df-zh-en}).
While this reinforces that models trained on filtered data outperform those trained on random samples, \mint{} makes it hard to understand which filtering metric is the best.

\begin{table*}[t]
    \begin{center}
        \setlength{\tabcolsep}{3pt}
\footnotesize
\begin{tabular}{l cc c@{\hspace{.4cm}} ccccccc}
    \toprule
    & \multicolumn{2}{c}{\textbf{Lexical}} & & \multicolumn{7}{c}{\textbf{Neural}} \\
    Filtering & \chrf{}$\uparrow$ & \bleu{}$\uparrow$ & & \comet{}$\uparrow$ & \bleurt{}$\uparrow$ & \cometkiwi{}$\uparrow$ & xC-REF$\uparrow$ & xC-QE$\uparrow$ & MX-REF$\downarrow$ & MX-QE$\downarrow$ \\
    \midrule
    Random & 24.36 & 21.70 & & 76.78 & 59.71 & 67.79 & 70.19 & 69.47 & 3.78 & 3.93 \\
    \cdashlinelr{1-11}
    \cometkiwi{} & \colorbox{green!20}{\textbf{26.57}} & \colorbox{green!20}{\textbf{26.16}} & & \colorbox{green!20}{\textbf{79.28}} & \colorbox{green!20}{\textbf{62.73}} & \colorbox{green!20}{\fbox{\textbf{72.99}}} & \colorbox{green!20}{74.35} & \colorbox{green!20}{73.23} & \colorbox{green!20}{2.86} & \colorbox{green!20}{2.59} \\
    \xcomet{} & \colorbox{green!20}{26.13} & \colorbox{green!20}{24.16} & & \colorbox{green!20}{77.51} & \colorbox{red!20}{59.05} & \colorbox{green!20}{68.54} & \colorbox{green!20}{\textbf{75.27}} & \colorbox{green!20}{\fbox{\textbf{74.75}}} & \colorbox{green!20}{2.97} & \colorbox{green!20}{3.30} \\
    \metricx{} & \colorbox{red!20}{24.04} & \colorbox{green!20}{23.07} & & \colorbox{green!20}{78.74} & \colorbox{green!20}{61.87} & \colorbox{green!20}{72.08} & \colorbox{green!20}{74.85} & \colorbox{green!20}{73.81} & \colorbox{green!20}{\textbf{2.66}} & \colorbox{green!20}{\fbox{\textbf{2.44}}} \\
    \bottomrule
\end{tabular}
    \end{center}
    \caption{System-level performance on the \wmttwothree{} English$\rightarrow{}$Chinese test set across several metrics of \gemma{}-2-2B with different training data filtering strategies.
    Cells where the interference and evaluation metrics are the same have \fbox{boxes} around values.
    Cells are painted \colorbox{green!20}{green} if they represent an improvement with respect to random filtering, \colorbox{red!20}{red} otherwise. Values in \textbf{bold} are the highest of each evlauation metric.
    }
    \label{tab:sys-level-filter}
\end{table*}

\subsection{Minimum Bayes Risk Decoding I: system-level analysis}\label{sec:analysis-mbr-i}

\paragraph{Interfering metrics prefer the systems they were used on and there are disagreements.}
Table~\ref{tab:sys-level-mbr} shows system-level evaluations on English$\rightarrow{}$Chinese for MBR.
Again, systems optimized for a certain metric score higher on that metric, with considerable differences in scores (e.g., >1 \comet{} point).\footnote{According to \citet{kocmi-etal-2024-navigating} this would mean that humans would prefer the winning system $>$90\% of the time; that percentage is likely overestimated due to \mint{}, though.}
However, unlike data filtering, there are significant disagreements among metrics.
With respect to greedy decoding, lexical metrics always deteriorate when performing MBR with neural metrics---as reported by~\citet{fernandes2022quality}.
Neural metrics can also disagree, like \bleurt{} and \comet{}, but this is an exception: every pair of neural metrics agree on all other LPs (see Tables~\ref{tab:apx-mbr-en-de}~to~\ref{tab:apx-mbr-zh-en}).
Conversely, performing MBR with lexical metrics rarely leads to improvements on any metric.
In this case, it is both hard to choose the best MBR metric, and whether improvements over greedy decoding are \textit{actually real}.

\iffalse
\paragraph{MBR with lexical metrics can lead to degradation across the board.}
%
Performing MBR with lexical metrics leads to performance degradations across the board, even according to \andre{the lexical metrics} themselves \andre{does this mean there is no \mint{} for lexical metrics?}.
%
This might be due to their nature: lexical metrics are not learned but rather depend solely on the similarity between the hypothesis and the reference translation.
%
If no candidate in the pool used for MBR is similar to the reference used in evaluation, there is no way the lexical metric can improve.
%
On the other hand, neural metrics are sensitive to changes in the hypothesis---and the source if they use it; this means there is room for optimization during MBR irrespective of the reference used for evaluation \andre{I don't understand this last point. is this relevant to be in the paragraph (which is about lexical metrics)?}
\fi

%\tocite{kinda obvious but citation? also could exaplain better.}

\begin{table*}[t]
    \begin{center}
        \setlength{\tabcolsep}{3pt}
\footnotesize
\begin{tabular}{l cc c@{\hspace{.4cm}} ccccccc}
    \toprule
    & \multicolumn{2}{c}{\textbf{Lexical}} & & \multicolumn{7}{c}{\textbf{Neural}} \\
    Decoding & \chrf{}$\uparrow$ & \bleu{}$\uparrow$ & & \comet{}$\uparrow$ & \bleurt{}$\uparrow$ & \cometkiwi{}$\uparrow$ & xC-REF$\uparrow$ & xC-QE$\uparrow$ & MX-REF$\downarrow$ & MX-QE$\downarrow$ \\
    \midrule
    Greedy & 43.23 & \textbf{44.46} & & 87.22 & 72.89 & 81.23 & 87.27 & 84.31 & 1.40 & 1.34 \\
    \cdashlinelr{1-11}
    \multicolumn{9}{l}{\small \textbf{MBR}} \\
    $\chrf{}_{L}$ & \colorbox{green!20}{\fbox{\textbf{43.45}}} & \colorbox{red!20}{43.97} & & \colorbox{red!20}{87.15} & \colorbox{red!20}{72.78} & \colorbox{red!20}{81.16} & \colorbox{red!20}{86.76} & \colorbox{red!20}{84.14} & \colorbox{red!20}{1.42} & \colorbox{green!20}{1.32} \\
    $\bleu{}_{L}$ & \colorbox{red!20}{42.99} & \colorbox{red!20}{\fbox{44.34}} & & \colorbox{red!20}{87.17} & \colorbox{red!20}{72.79} & \colorbox{red!20}{81.08} & \colorbox{red!20}{86.82} & \colorbox{red!20}{84.02} & \colorbox{red!20}{1.43} & \colorbox{red!20}{1.35} \\
    \cdashlinelr{1-11}
    $\comet{}_{N}$ & \colorbox{red!20}{39.75} & \colorbox{red!20}{40.21} & & \colorbox{green!20}{\fbox{\textbf{88.38}}} & \colorbox{red!20}{72.55} & \colorbox{green!20}{\textbf{81.95}} & \colorbox{green!20}{\textbf{87.95}} & \colorbox{green!20}{\textbf{85.58}} & \colorbox{green!20}{1.30} & \colorbox{green!20}{\textbf{1.17}} \\
    $\bleurt{}_{N}$ & \colorbox{red!20}{38.91} & \colorbox{red!20}{39.14} & & \colorbox{red!20}{87.06} & \colorbox{green!20}{\fbox{\textbf{74.33}}} & \colorbox{green!20}{81.65} & \colorbox{green!20}{87.61} & \colorbox{green!20}{85.41} & \colorbox{green!20}{\textbf{1.29}} & \colorbox{green!20}{1.18} \\
    \bottomrule
\end{tabular}
    \end{center}
    \caption{System-level performance on the \wmttwothree{} English$\rightarrow{}$Chinese test set across several metrics of \Towervtwo{}-7B with different decoding strategies.
    Cells where the interference and evaluation metrics are the same have \fbox{boxes} around values.
    Cells are colored \colorbox{green!20}{green} if they represent an improvement with respect to greedy, \colorbox{red!20}{red} otherwise. Values in \textbf{bold} are the highest of each evlauation metric.
    Subscripts indicate metric family: L means lexical, N is Neural.
    }
    \label{tab:sys-level-mbr}
\end{table*}

\paragraph{LLM Judges also suffer from \mint{}.}
Table~\ref{tab:reduced-results-judges} shows additional results using LLM Judges \prometheus{} and \iscore{} for evaluation and MBR on Chinese$\rightarrow$English.\footnote{Out of the \wmttwothree{} LPs, \iscore{} only supports Chinese$\rightarrow$English and English$\rightarrow$German.} %, which we leave in Appendix~\ref{apx:other-lps}}
Both judges behave similarly to neural metrics in that they improve according to themselves and cause degradation on all MT-specific metrics.
%
%\iscore{}---an MT-specific judge---performs slightly better according to other metrics.
%, this shows there is significant room for improvement in LLM Judges for MT.

\begin{table}[t]
    \centering
\footnotesize
\begin{tabular}{lcccc}
    \toprule
    Decoding & \chrf{}$\uparrow$ & \comet{}$\uparrow$ & Ptheus$\uparrow$ & IScore$\uparrow$ \\
    \midrule
    Greedy & \textbf{49.80} & 81.58 & 3.54 & -6.07\\
    \cdashlinelr{1-5}
    \comet{} & \colorbox{red!20}{48.87} & \colorbox{green!20}{\fbox{\textbf{82.32}}} & \colorbox{green!20}{3.58} & \colorbox{green!20}{-5.83} \\
    Ptheus & \colorbox{red!20}{46.99} & \colorbox{red!20}{80.78} & \colorbox{green!20}{\fbox{\textbf{4.06}}} & \colorbox{green!20}{-5.99} \\
    IScore & \colorbox{red!20}{47.24} & \colorbox{red!20}{81.00} & \colorbox{green!20}{3.66} & \colorbox{green!20}{\fbox{\textbf{-4.86}}} \\
    \bottomrule
\end{tabular}
    \caption{Variant of Table~\ref{tab:sys-level-mbr} for Chinese$\rightarrow$English with two LLM Judges: \prometheus{} 7B (Ptheus) and \iscore{} (IScore). Values in \textbf{bold} are the highest of each evlauation metric.}
    \label{tab:reduced-results-judges}
\end{table}

%\subsection{Impact of Metric Interference on Metrics of Different Types}

%\paragraph{Neural metrics increase under neural metric interference while lexical metrics decrease.}
%When the interfering metric is neural, other neural metrics also increase.
%
%This is expected, given the high correlations among them (see~\ref{fig:corr-matrix}), but also highlights the issue of performing evaluation neural metrics other than the interference one.
%
%On the other hand, as reported by~\citet{fernandes2022quality}, we observe that lexical metrics decrease   after MBR with neural metrics.
%
%Most works report gains in human terms from MBR, so a part of this difference must be due harmless lexical differences between model generations and references.
%
%However, this deterioration in lexical metrics may still contain valuable information regarding over-optimization of the interfering metric.

%\paragraph{Lexical metrics can improve with data filtering.}
%While MBR seems negative for lexical metrics in general, that is not the case with data filtering.
%
%Both \bleu{} and \chrf{} improve when filtering data with \cometkiwi{} and \xcomet{}.
%
%This is more evidence that data filtering is a ``softer'' case of \mint{}, which has fewer harmful consequences for automatic evaluation.

\subsection{Minimum Bayes Risk Decoding II: instance-level analysis}\label{sec:analysis-mbr-ii}
We performed an instance-level analysis to understand the root cause of the system-level findings in Section~\ref{sec:analysis-mbr-i}.
%
%Namely, the disagreement between neural and lexical metrics, and the correlations among neural metrics.
%
We focused on comparing differences in metric scores between pairs of models with disparate performances.
One pair is \Towervtwo{}-70B and \Towervtwo{}-7B with greedy decoding, which represents a metric-agnostic improvement.
The other is \Towervtwo{}-7B with \comet{} MBR and greedy decoding, an improvement driven by \mint{}.
Comparing the two pairs allows us to isolate the effect of \mint{} on metric score deltas to better understand metric disagreements.
%This setup is similar in spirit to difference-in-differences~\citep{ashenfelter1984using} \andre{can you explain what this is?}: the treatment is MBR, and the control and treatment groups are the first and second model pairs, respectively.

\begin{figure}[t]
    \centering
    \includegraphics[width=\linewidth]{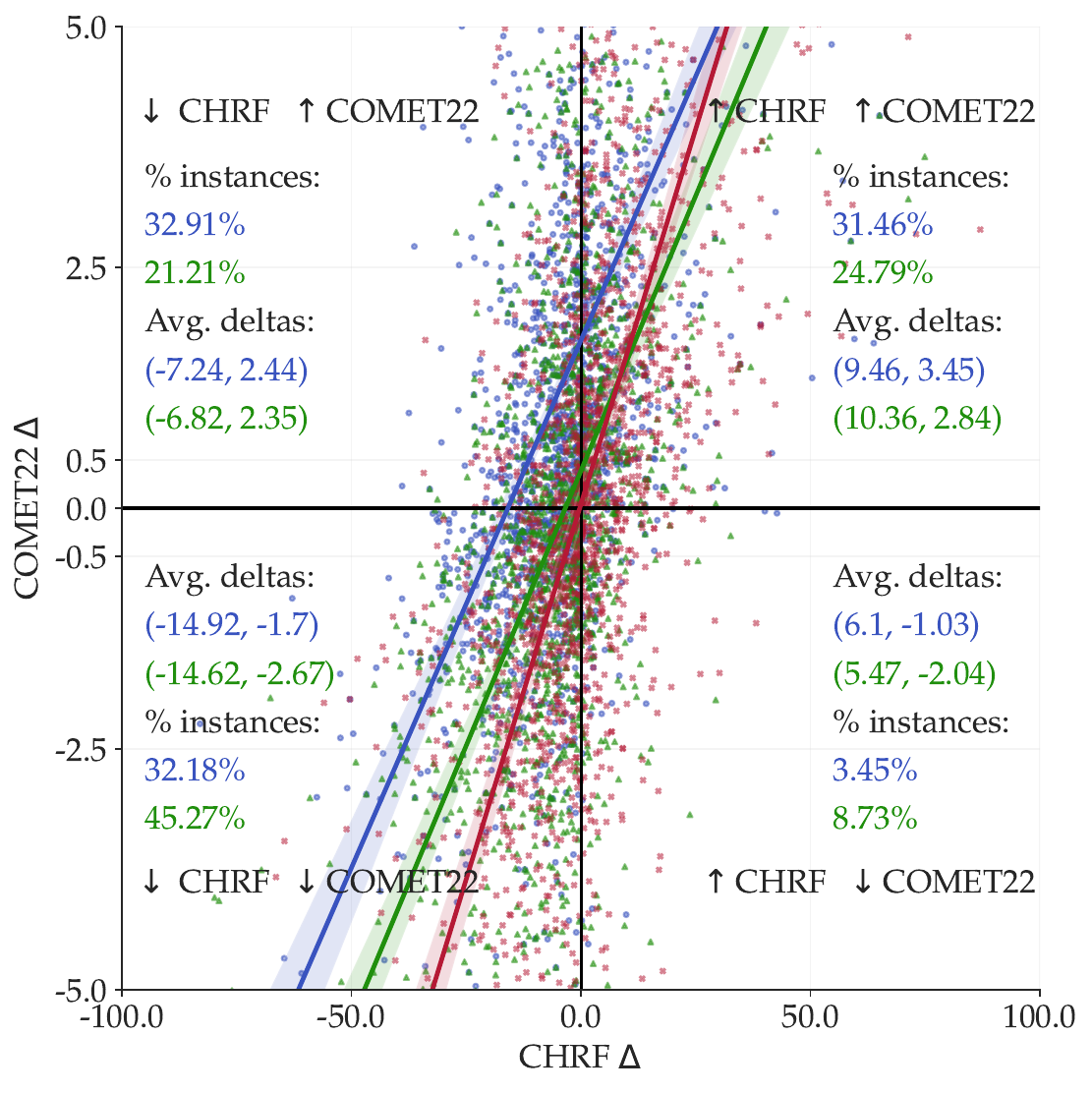}
    \caption{Difference ($\Delta$) in \comet{} and \chrf{} between models of three sets: (\textcolor{CustomRed}{\Towervtwo{}-70B greedy, \Towervtwo{}-7B greedy}); (\textcolor{CustomBlue}{\Towervtwo{}-7B + \comet{} MBR, \Towervtwo{}-7B greedy}); (\textcolor{CustomGreen}{\Towervtwo{}-7B + \bleurt{} MBR, \Towervtwo{}-7B greedy}).
    We include a linear regression for each set of points. 
    Metric interference causes distortions in score relationships between the interfering metric and correlated metrics.
    %Notice that the MBR regressions do not cross the origin, meaning that, in both cases, a change of 0 in \chrf{} corresponds to an increase in \comet{}, on average (an indication of \comet{} over-optimization).
    }
    \label{fig:diff-scatter}
\end{figure}

\iffalse
\paragraph{The interfering metric increases on the majority of instances relative to the baseline.}
System-level results provide a useful yet limited view of performance because of aggregation.
%
They do not reveal whether improvements are driven by a minority of instance that improve a lot in quality, or if they are consistent across most instances.
%
This kind of information is important to understand better how \mint{} can be mitigated.
%
Figure~\ref{fig:diff-scatter} shows that the majority of instances improve in quality after MBR but the quality of a considerable amount of instances~(roughly 35\%) degrades.\footnote{Plots for other LPs are in Appendix~\ref{apx:other-lps} and yield similar findings.}
%
This finding is somewhat unexpected, given the system-level improvements we reported earlier---\mint{} impacts instances in distinct, non-trivial ways.
\fi

\paragraph{System-level metric disagreement is \textit{actually} driven by a minority of instances.}
Figure~\ref{fig:diff-scatter} shows the difference in \chrf{} and \comet{} between the aforementioned model pairs (in red and blue, respecitvely) for English$\rightarrow$Chinese.
When comparing models where \mint{} is not an issue, we see that most instances ($>$70\%) lie in quadrants where both metrics agree on the direction of change.
When comparing models under \mint{}, a mass of instances shifts towards the disagreement quadrant that favors the interfering metric (\comet{}).
The shift is so pronounced that a delta of 0 points in \chrf{} corresponds to a non-zero change in \comet{},\footnote{The change is almost 2 points on average, which, according to~\citet{kocmi-etal-2024-navigating}, would correspond to humans preferring translations from the system 95\% of the time in a setting without~\mint{}.
Surely, given the observed score distortions, we cannot make such interpretations under \mint{}.} highlighting a stark distortion in the scores of the interfering metric.
That said, this only occurs to roughly 33\% of instances; for most instances (around 63\%), metrics still agree on the direction of change.
Similar phenomena occur among other metrics (we omit these plots for the sake of brevity).
%
%This finding is not fully in line with the  system-level disagreements we reported earlier, further suggesting that instance-level information is crucial for understanding \mint{}.

\paragraph{Metric interference score distortions affect correlated metrics.}
A common strategy to mitigate \mint{} is to use different metrics for interference and evaluation.
Figure~\ref{fig:diff-scatter} shows a third model pair in green: \Towervtwo{}-7B with \bleurt{} MBR and greedy decoding.
The impact of \bleurt{} MBR in distorting metric scores is similar to that of \comet{} MBR, though less pronounced.
This is likely because \bleurt{} is highly correlated with \comet{},\footnote{It is generally known that neural metrics correlate strongly due to similar training data and architectures~\citep{kocmi-etal-2024-navigating,kovacs-etal-2024-mitigating}. We include an instance-level correlation matrix in Table~\ref{tab:apx-corr-matrix} to further support this; \comet{}-\bleurt{} correlation is 0.75.
Furthermore, on system level evaluation, while the two metrics disagree when using the other for MBR on English$\rightarrow$Chinese, they agree on all other LPs (see Tables~\ref{tab:apx-mbr-en-de} to~\ref{tab:apx-mbr-zh-en}. 
If anything, the result on English$\rightarrow$Chinese shows how system-level information is insufficient to fully understand \mint{}.} %, highlighting the importance of instance-level analysis.} 
and it highlights the inadequacy of using correlated metrics for evaluation under \mint{}: they may be less biased than the interfering metric but they will still be biased. 
%\andre{I'm a bit confused here. Table 3 earlier seems to suggest that at system level improvements in COMET (when doing MBR with COMET) correspond to degradations in BLEURT, and vice-versa. doesn't that suggest the BLEURT and COMET are not so correlated after all? or is this something that happens at system level but does not seem to hold for many instances?}\pombal{Expanded footnote 9 with this. At a system-level comet and bleurt agree on all LPs except for english-chinese (table 3). I thought about changing the LP in the main text but I like English-Chinese since all phenomena are more noticeable. This ``inconsistency'' can even be a good thing: the instance-level analysis we make shows something \textit{weird} is happening, even in cases where metrics don't seem affected at a system level.}

\begin{figure}[t]
    \centering
    \includegraphics[width=\linewidth]{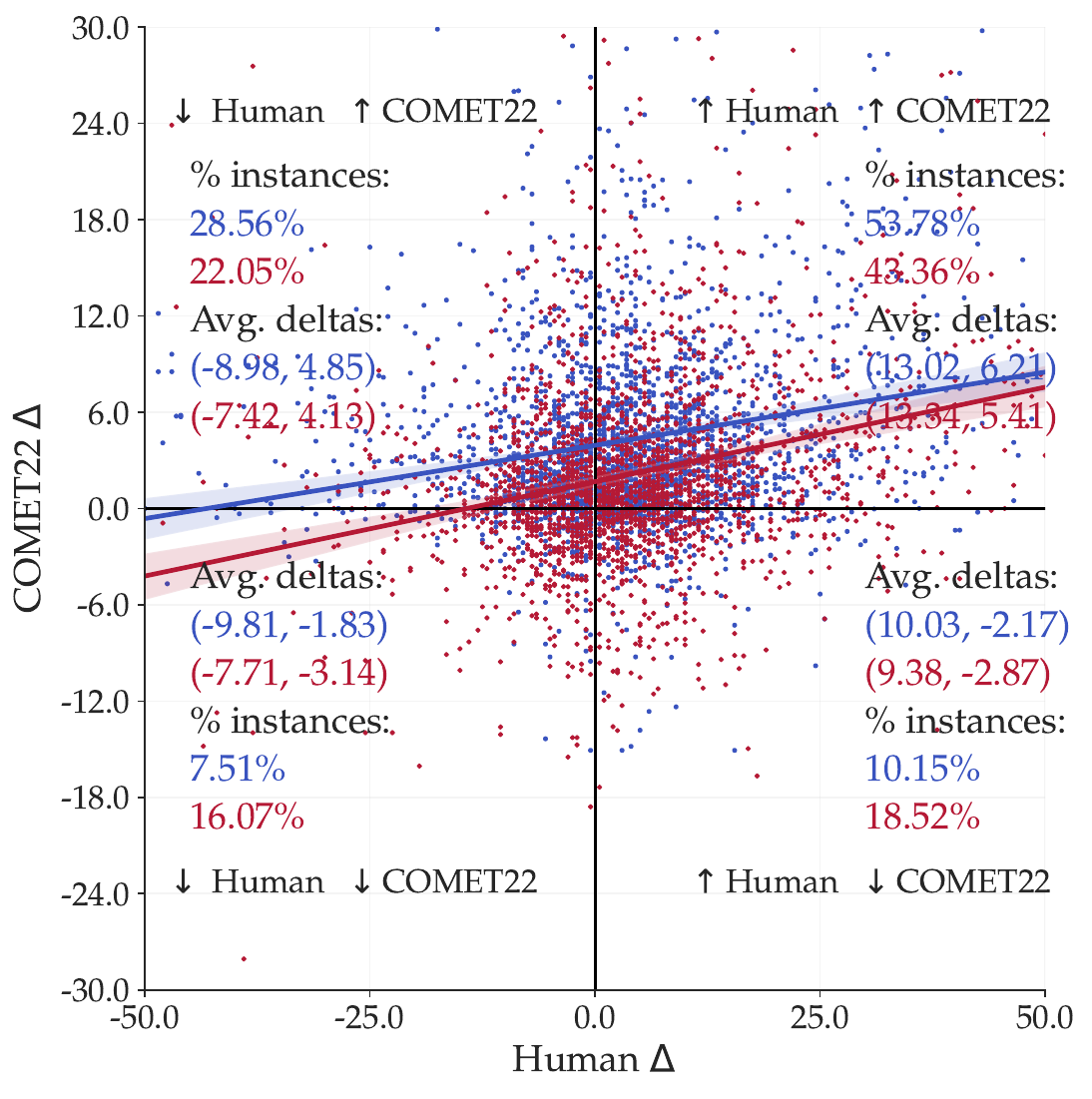}
    \caption{Difference ($\Delta$) in \comet{} and Human scores on \wmttwofour{} English$\rightarrow$Chinese of two sets of systems (\mint{} affects the blue set).
    %
    % consider putting more points in the human scatter, or show more scatters.
    %We observe similar score distortions to Figure~\ref{fig:diff-scatter}, which compares automatic metrics.
    }
    \label{fig:diff-scatter-wmt24}
\end{figure}

\paragraph{Instance-level metric distortions extend to human evaluation.}
Figure~\ref{fig:diff-scatter-wmt24} shows that the same score distortions we observed among automatic metrics hold between an interfering metric and human scores.
We use \wmttwofour{} English$\rightarrow$Chinese data because it has both human evaluations and systems that are comparable to our initial setup. 
We consider system sets instead of pairs for a larger sample of points.
Our representative unbiased system (previously \Towervtwo{}-70B greedy) becomes \gptfour{}, the best model in this LP according to human evaluation.
The biased system is still \Towervtwo{}-70B with \comet{} MBR.
For each translation of these systems, we obtain deltas from four other systems that we are sure did not leverage \comet{} (yielding four points of each color per translation): IOL-Research, IKUN, IKUN-C, and Llama-3-70B.
%
%Similar phenomena hold for other system sets.
%
%This parallel further hints at the usefulness of instance-level information for tackling \mint{}.

\subsection{Final Remarks} 
We analyzed the impact of \mint{} on evaluation with several metrics at system- and instance-levels.
Our system-level findings corroborate those of existing literature, and our instance-level analysis adds an important layer of understanding: \mint{} causes non-trivial score distortions between the interfering metric and other metrics and human scores.
Importantly, these distortions extend to correlated metrics.
These findings lead to two conclusions: 1) instance-level information is crucial for understanding \mint{}; 2) performing evaluation with metrics that are spuriously correlated with the interfering metric is potentially just as misleading as using the interfering metric itself, bringing common practices into question.
%
% The case of \Towervtwo{} in \wmttwofour{} illustrates the latter: even though MBR was performed with \comet{}, the model's performance with \metricx{} and \cometkiwi{}-XXL was also significantly overestimated.
%
%In the next section, we propose a new method for evaluation under \mint{} which leverages these insights by combining instance-level information from several metrics.

\section{Improving Evaluation Under \mint{}}\label{sec:adjusted-metric}
%The previous section highlighted the need for developing improved evaluation practices in cases of \mint{}.
%
%Addressing this open problem may involve, for example, correcting existing metrics or creating new ones.
%
%In this section, we focus on adjusting the scores of the interfering metric. 
%
%We understand that practitioners often prefer using familiar metrics, and this avoids introducing a new one that may be harder to interpret.
%
%Existing metrics like \comet{} and \bleu{} have been the subject of extensive work on improving interpretability, such as the connection between metric improvements and human preference demonstrated by \citet{kocmi-etal-2024-navigating} -- insights we build upon in this work.

\subsection{\methodname{}}
Leveraging the insights of Section~\ref{sec:analysis}, we propose \methodname{}, a method for correcting the instance scores of the interfering metric during evaluation (see Figure~\ref{fig:method-fig}). %, making it more reliable.
\methodname{} does not require human judgements, instead leveraging the scores of other metrics. % -- the closest proxy for human judgements available.
For a test set $\mathcal{D}$ containing (source, translation, reference) triplets $\langle x, h, r \rangle$, \methodname{} learns to predict the interfering metric, $\hat{q}_{\text{int}}$, using a set of other, less biased metrics, $\{\hat{q}_{\text{safe}}\}$, as features.
\methodname{} is trained on translations from one or more models that are not subject to \mint{}, and predicts the $\hat{q}_{\text{int}}$ scores of a model affected by \mint{}, producing $\hat{q}_{\text{adjusted}}$.
The framework is agnostic to the regression method.
%
%Figure~\ref{fig:method-fig} summarizes how \methodname{} works.
%(e.g., linear regression, random forest, etc.)

\methodname{} is inspired by the analysis performed in Section~\ref{sec:analysis}: we found that \mint{} distorts the relationships of instance-level scores among metrics; thus, we hypothesize that a less biased version of the interfering metric can be learned from the instance-level scores of other metrics on models where \mint{} is not a concern.
It should be noted that \mint{} distorts all spuriously correlated metrics, not just the interfering one, so the inputs of \methodname{} at inference time will be biased.
However, the distortion is less pronounced, and by leveraging a large set of metrics we posit the biases of each one may be diluted.
Next, we detail the experimental setup for evaluating \methodname{}.

\subsection{Experimental Setup}
\paragraph{Overview.} Meta-evaluating (i.e., evaluating an evaluation method) \methodname{} requires: 1) translations of one or more systems that did not leverage metrics; 2) translations of one or more systems that used metrics (\mint{}); 3) knowledge of the interfering metric; 4) human judgments for the translations.
%\andre{maybe we could have a peppermint emoji for \mint{} which we could use throughout the paper to mark models with \mint{}?}
%
The most recent edition of the \wmttwofour{} MT shared task~\citep{kocmi2024findings} contains all this information for state-of-the-art MT systems on 11 language pairs.\footnote{The language pairs are English-German, -Icelandic, -Czech, -Spanish, -Ukrainian, -Russian, -Japanese, -Chinese, -Hindi, Czech-Ukrainian, and Japanese-Chinese. The amount of evaluated systems per language pair ranges from 11 to 19.} Furthermore, one relevant consequence of \mint{} in this shared task was \Towervtwo{}-MBR---which used \comet{} for MBR---ranking first on all language pairs according to automatic metrics but not to humans.

\paragraph{Meta-evaluation metrics.}
%\footnote{\Towervtwo{} was the best submission on 8 of 11 LPs but only ranked first on 1 after considering all baselines (e.g., \claudethreefive{}, \gptfour{}, etc.)}
%
The goal of \mint{} mitigation methods is to rank the model affected by~\mint{} more accurately.
Thus, for system pairs that contain \Towervtwo{}, we report \textbf{soft pairwise accuracy (SPA)}~\citep{thompson2024improving}---the system-level meta-metric used in the latest WMT metrics shared task~\citep{freitag2024llms}:
\begin{equation}\label{eq:spa}
    SPA = \binom{N}{2}^{-1} \sum_{i=0}^{N-1} \sum_{j=i+1}^{N-1} 1 - \left| p_{ij}^h - p_{ij}^m \right|
\end{equation}
N is the number of systems, $p^{h}_{ij}$ the p-value for the hypothesis that humans prefer system $i$ over $j$, and $p^{m}_{ij}$ is the p-value for the hypothesis that a metric prefers system $i$ over $j$.
%
%The combinatorial term normalizes by the total number of unique system pairs being compared.
%
SPA judges metrics both on whether they rank systems like humans and on whether their confidence in the rankings is similar.
%
%An added benefit of SPA is that ties are not an issue because comparisons are not binary.

Since instance-level meta-evaluation is also important to discriminate between metrics~\citep{freitag2022results}, we also report an \textbf{instance-level version of pairwise accuracy (Ins-PA)}:
%\andre{this needs to be better explained, the notation $\mbox{metric} \Delta$ is not explained and you need to explain why using the sign. I understand this is about flipping the rank, but might not be clearer to all readers. maybe there is a better notation, e.g. the kind of notation used in preference ranking, e.g. $\frac{|\{(x, h_+, h_-, r) \in \mathcal{D} \,:\, \mathcal{M}(x, h_+, r) > \mathcal{M}(x, h_-, r)\}|}{|\mathcal{D}|}$}
\begin{equation}\label{eq:ins-pa}
    \text{Ins-PA} = \frac{\sum_{d \in \mathcal{D}} \mathbb{I}\left[ \mathcal{M}(d_+) > \mathcal{M}(d_-) \right]}{\mathcal{|D|}}
\end{equation}
where $\mathcal{D}$ is the set, across all system pairs, of all unique tuples, $d$, containing a source, its human-favored translation, its human-disfavored translation, and a reference ($\langle x, h_+, h_-, r\rangle$); $d_+$ corresponds to $\langle x, h_+, r\rangle$, $d_-$ corresponds to $\langle x, h_-, r\rangle$, and $\mathcal{M}$ is a metric's judgement. If the human score for $d_+$ and $d_-$ is the same, that instance is scored 1 only if $\mathcal{M}(d_+) = \mathcal{M}(d_-)$.\footnote{As such, while Ins-PA considers ties, it does not calibrate a threshold for ties, as the meta-metric used in the metrics shared task~\citep{deutsch2023ties}. 
If anything, this will penalize \methodname{} and favor \xcomet{} or \metricx{}, which produce more ties for perfect translations.
We computed results when disregarding ties and reached similar conclusions.
}

\paragraph{Further meta-evaluation details.}
When aggregating the aforementioned metrics across LPs, we report their macro-average, and, for Table~\ref{tab:metric-adjust-sys-ins}, a normalized Borda count~\citep{colombo2022best} by averaging the ranks of each metric over all LPs (lower is better).
This metric is better at representing consistency across LPs than an average, which is more sensitive to outliers.
Furthermore, since metrics are known to perform differently when comparing low- and high-quality systems, we also report results on a subset of high-quality systems: namely, \Towervtwo{} pairs with \gptfour{}, \claudethreefive{}, and \gemini{}, when available.\footnote{\gemini{} is not available for Icelandic.}
%
%This focuses on how good metrics are for establishing the state-of-the-art.
%
%This is a particularly relevant comparison when developing state-of-the-art models.

\paragraph{Metric baselines.}
As mentioned before, the usual practice under \mint{} is to either keep using the interfering metric for evaluation, or to use a single other metric.
Thus, our baselines are \comet{} and representative state-of-the-art MT metrics of several families: \chrf{} (lexical), \cometkiwi{} (reference-less), \xcomet{}-XL, and \metricx{}-XL (neural).

\paragraph{Ensemble baselines.}
We report two baselines that involve ensembling metric scores: \autorank{}~\citep{kocmi2024findings} and \metametric{}~\citep{anugraha-etal-2024-metametrics}.
The former was used to obtain the preliminary rankings of the \wmttwofour{} MT shared task, and the latter is a state-of-the-art metric ensembling method, according to the results of the \wmttwofour{} shared task.
\autorank{} works by, for each LP, linearly scaling (between 1 and the number of systems) a set of system-level metric averages and then computing the average among the metrics to obtain a single value per system.
In the shared task, \metricx{}-XL and \cometkiwi{}-XL were used; for the sake of comparability with our setup, we ensemble the baseline metrics stated in the previous paragraph, except for \comet{}.
\metametric{}, on the other hand, like most other metric, learns to predict translation scores of humans.
However, instead of leveraging vectorial representations of text as learning features, it uses the scores of other metrics, therefore functioning as a learned ensemble.
The authors use Gaussian Processes to this end, but \metametric{} is agnostic to the regressor class.
For the sake of comparability, we use the same regression model as \methodname{} (the Random Forest described in the ``Training details'' paragraph of this section), and the same metrics for ensembling as \autorank{}.
Crucially, \metametric{} requires an annotated training set, while \methodname{} does not.
We use all the human assessments available for all language pairs of \wmttwothree{} for training.\footnote{This includes 9 language pairs: Czech$\rightarrow$Ukrainian, English$\rightarrow$\{Czech, German, Japanese, Chinese\}, and \{German, Hebrew, Japanese, Chinese\}$\rightarrow$English. Training a single regression model on all LPs (the alternative suggested by~\citet{anugraha-etal-2024-metametrics} for whenever a test LP is not seen in training) yielded better results than training LP-specific models.}

\paragraph{\autorank{}-Ins baseline.}

As an additional baseline, we propose a instance-level version of \autorank{}.
Given an ensemble of metrics, for each LP, \autorank{}-Ins is obtained by linearly scaling the score of each metric between 1 and the total number of instances (across all systems), and then averaging this over all the metrics in the ensemble.
\autorank{}-Ins yields instance-level scores that can be averaged to obtain a system-level scores.

\paragraph{Training details.} We use \methodname{} to predict \comet{} scores for \Towervtwo{}-MBR on all language pairs of the \wmttwofour{} test set.
%a submission to the shared task that leveraged \comet{} MBR decoding with \Towervtwo{}.
%  \andre{``combined'' is probably not the best word, I suggest you mention here right away \Towervtwo{} and \Towervtwo{}-MBR and then say you use \methodname{} to debias the score of \Towervtwo{}-MBR} with MBR decoding with \comet{}
%
In the main results, for training, we use scores of baseline metrics from the previous paragraph , except for \comet{}, on \Towervtwo{} greedy \wmttwofour{} translations as features (we explore other training metric features and systems later).
For each language pair, we train a regressor to predict \comet{} scores.
This trained regressor is \methodname{}.
At inference time, we query \methodname{} for every instance of every system. %, not just \Towervtwo{}-MBR.
Our regressor is a random forest with 1000 trees, a maximum depth of 4, and the default hyperparameters of \texttt{scikit-learn}~\cite{pedregosa2011scikit}, which we detail in Appendix~\ref{apx:rf-hp}.\footnote{We do not optimize hyperparameters, and opt for a relatively high number of estimators but low maximum depth to avoid overfitting.}

\iffalse
\begin{table*}[h]
    \begin{center}
        \include{tables/metric_adjust_sys}
    \end{center}
    \caption{System-level pairwise accuracy for baseline metrics and \methodname{} variants (training translations) on the \wmttwofour{} test set.
    \comet{} (with the *) is the interference (and adjusted) metric.
    \textbf{Bold} values are the best overall; values in \textit{italic} are cases where \methodname{} improves upon the interfering metric (\comet{}); \underline{underlined} values are cases where \methodname{} beats other baselines. Ties between our method and baselines are not marked.
    }
    \label{tab:metric-adjust-sys}
\end{table*}

\begin{table*}[h]
    \begin{center}
        \include{tables/metric_adjust_ins_claude}
    \end{center}
    \caption{Instance-level pairwise accuracy for baseline metrics and \methodname{} variants (training translations) on the \wmttwofour{} test set between \Towervtwo{} and \claudethreefive{}.
    \comet{} (with the *) is the interference (and adjusted) metric.
    \textbf{Bold} values are the best overall; values in \textit{italic} are cases where \methodname{} improves upon the interfering metric (\comet{}); \underline{underlined} values are cases where \methodname{} beats other baselines. Ties between our method and baselines are not marked.
    }
    \label{tab:metric-adjust-ins-claude}
\end{table*}
\fi

\subsection{Results \& Discussion}

\begin{table*}[t]
    \begin{center}
\renewcommand{\arraystretch}{1.2}
\footnotesize
\begin{tabular}{l cccc cccc}
    \toprule
    & \multicolumn{4}{c}{\textbf{System-level}} & \multicolumn{4}{c}{\textbf{Segment-level}} \\
    \cmidrule(r){2-5} \cmidrule(l){6-9}
    & \multicolumn{2}{c}{\textbf{Average $\uparrow$}} & \multicolumn{2}{c}{\textbf{Borda Count $\downarrow$}} & \multicolumn{2}{c}{\textbf{Average $\uparrow$}} & \multicolumn{2}{c}{\textbf{Borda Count $\downarrow$}} \\
    Metrics & All & High-Q & All & High-Q & All & High-Q & All & High-Q \\
    \midrule
    \multicolumn{9}{l}{\small \bf Baselines} \\
    \multicolumn{9}{l}{\small \bf Metrics} \\
    $\chrf{}_{L,REF}$ & 0.5958 & 0.4556 & 5.4545 & 5.4545 & 0.4548 & 0.4328 & 6.3636 & 5.0000 \\
    $\cometkiwi{}_{N,QE}$ & 0.8219 & 0.6476 & 3.0909 & 2.6364 & 0.4716 & 0.4345 & 5.3636 & 5.1818 \\
    $\xcomet{}_{N,REF}$ & 0.8199 & 0.6476 & 3.2727 & 2.6364 & 0.4968 & 0.4510 & 3.1818 & 3.3636 \\
    $\metricx{}_{N,REF}$ & 0.8219 & 0.6522 & 2.5455 & 2.3636 & 0.4868 & 0.4385 & 4.4545 & 4.4545 \\
    $\comet{}_{N,REF}*$ & 0.8277 & 0.6775 & 3.0000 & 2.1818 & 0.4955 & 0.4458 & 3.0000 & 3.4545 \\
    \multicolumn{9}{l}{\small \bf Ensembles} \\
    $\autorank{}_{E}$ & 0.8199 & 0.6476 & 3.2727 & 2.6364 & -- & -- & -- & -- \\
    $\metametric{}_{E}$ & 0.8080 & 0.6506 & 4.1818 & 2.4545 & 0.4434 & 0.4115 & 7.4545 & 6.5455 \\
    \cdashlinelr{1-9}
    \multicolumn{9}{l}{\small \bf This work} \\
    \autorank{}-Ins$_{E}$ & 0.8190 & 0.6757 & 3.0000 & 2.0909 & \textbf{0.4978} & 0.4509 & \textbf{2.3636} & 2.9091 \\
    $\methodname{}_{E}$ & \textbf{0.8278} & \textbf{0.6846} & \textbf{2.4545} & \textbf{1.5455} & 0.4956 & \textbf{0.4551} & 3.2727 & \textbf{2.3636} \\
    \bottomrule
\end{tabular}
    \end{center}
    \caption{System-level SPA and instance-level PA for baseline metrics, \autorank{}-Ins, and \methodname{} on the \wmttwofour{} test set for two sets of system pairs: all pairs that contain \Towervtwo{}-MBR (All), and a subset of high-quality system pairs containing \Towervtwo{}-MBR (High-Q).
    \comet{} (with the *) is the interference (and adjusted) metric.
    \textbf{Bold} values are the best overall.
    Subscripts indicate metric family: L means lexical, N is Neural, QE is reference-less, REF is reference-based and E is ensemble.%\pombal{still considering showing only borda...}
    %\pombal{I wonder if we should only keep Borda count and point to LP + avg results in the Appendix.}
    }
    \label{tab:metric-adjust-sys-ins}
\end{table*}

\begin{table*}[t]
    \begin{center}
    \setlength{\tabcolsep}{3pt}
\renewcommand{\arraystretch}{1.3}
\footnotesize
\begin{tabular}{l ccccccccccc}
    \toprule
    Metrics & de & es & cs & ru & uk & is & ja & zh & hi & cs$\rightarrow{}$uk & ja$\rightarrow{}$zh \\
    \midrule
    \multicolumn{9}{l}{\footnotesize \bf Baselines} \\
    $\chrf{}_{L,REF}$ & 0.4338 & 0.5248 & 0.3069 & 0.4963 & 0.4010 & \textbf{0.8091} & \textbf{0.7655} & 0.5427 & 0.5790 & \textbf{0.9113} & \textbf{0.7835} \\
    $\cometkiwi{}_{N,QE}$ & 0.8912 & 0.6940 & 0.9983 & 0.9161 & 0.8266 & 0.7781 & 0.5697 & 0.9017 & 0.9161 & 0.8033 & 0.7457 \\
    $\xcomet{}_{N,REF}$ & 0.8912 & 0.6940 & 0.9983 & 0.9161 & 0.8266 & 0.7781 & 0.5477 & 0.9017 & 0.9161 & 0.8033 & 0.7457 \\
    $\metricx{}_{N,REF}$ & 0.8921 & 0.6940 & 0.9983 & 0.9161 & 0.8266 & 0.7781 & 0.5477 & 0.9017 & 0.9232 & 0.8132 & 0.7497 \\
    $\comet{}_{N,REF}*$ & 0.8912 & 0.6941 & 0.9983 & 0.9161 & 0.8138 & 0.7781 & 0.5477 & 0.9017 & 0.9162 & 0.8999 & 0.7472 \\
    $\autorank{}_{E}$ & 0.8912 & 0.6963 & 0.9983 & 0.9161 & 0.7156 & 0.7781 & 0.5489 & 0.9017 & 0.9227 & 0.8936 & 0.7465 \\
    $\metametric{}_{E}$ & 0.7514 & 0.6938 & 0.9983 & 0.9161 & 0.8266 & 0.7781 & 0.5477 & \textbf{0.9085} & 0.9159 & 0.8033 & 0.7485 \\
    \cdashlinelr{1-12}
    \multicolumn{12}{l}{\footnotesize \bf This work} \\
    \autorank{}-Ins$_{E}$ & 0.8912 & 0.6940 & 0.9983 & 0.9161 & \textbf{0.8266} & 0.7781 & 0.5477 & 0.9017 & 0.9161 & 0.8033 & 0.7457 \\
    $\methodname{}_{E}$ & \textbf{0.8937} & \textbf{0.7061} & \textbf{0.9983} & \textbf{0.9163} & 0.7154 & 0.7781 & 0.6023 & 0.8995 & \textbf{0.9346} & 0.9032 & 0.7582 \\
    \bottomrule
\end{tabular}
    \end{center}
    \caption{System-level SPA results for all LPs (direction omitted for from-English LPs).}
    \label{tab:metric-adjust-lp-sys-all}
\end{table*}

\paragraph{\methodname{} ranks systems and instances more accurately than metrics and \autorank{}.}
Table~\ref{tab:metric-adjust-sys-ins} shows our main system- and instance-level accuracy results.\footnote{We present per-LP results in Tables~\ref{tab:metric-adjust-lp-sys-all} and~\ref{tab:apx-metric-adjust-lp-seg-high-q}.}
At a system level, \methodname{} ranks systems more accurately on average and in terms of Borda count (meaning it is more consistent across LPs).
The superiority of \methodname{} is more noticeable when ranking high-quality systems; this comparison is particularly important for determining the \textit{de facto} state-of-the-art systems.
Conclusions are similar for the instance-level evaluation, where \autorank{}-Ins performs slightly better when considering all systems.
Additionally, Table~\ref{tab:metric-adjust-lp-sys-all} shows that, out of 11 LPs, \methodname{} outperforms all baselines on 5 LPs, ranks second on 4, and third on 2, highlighting the method's generalizability across languages.

\paragraph{\methodname{} outperforms \metametric{}, which requires annotated data.} Table~\ref{tab:metric-adjust-sys-ins} also shows that \methodname{} outperforms \metametric{} across the board.
This is an encouraging result, considering that \metametric{} requires human annotations.
Relying on human annotations can be especially problematic if the test set contains language pairs for which there is no pre-existing representative data, as is the case for Icelandic and Hindi, where \metametric{} performs particularly poorly (see Tables~\ref{tab:apx-metric-adjust-lp-seg-all} and \ref{tab:apx-metric-adjust-lp-sys-high-q} in the Appendix).
Even when data is available, it might not be representative, and building a training corpus still entails a series of design decisions.
\methodname{}, on the other hand, does not require human annotations as it is meant to be used directly on the test set of interest.
This finding hints at the benefits of learning to adjust a metric rather than ensembling metrics to learn human scores in cases of \mint{}.

\begin{table*}[t]
    \begin{center}\setlength{\tabcolsep}{3pt}
\renewcommand{\arraystretch}{1.2}
\footnotesize
\begin{tabular}{l cccccc}
    \toprule
    & \multicolumn{6}{c}{\textbf{Ensemble Metrics}} \\
    Method & Table~\ref{tab:metric-adjust-sys-ins} & All other metrics & QE only & REF only & Neural only & Lexical only \\
    \midrule
    \multicolumn{7}{l}{\textbf{\autorank{}}} \\
    \autorank{} & \colorbox{gray!20}{0.8199} & 0.8199 & 0.8199 & \colorbox{green!20}{0.8340} & 0.8199 & \colorbox{red!20}{0.5488} \\
    \autorank{}-Ins & \colorbox{gray!20}{0.8190} & 0.8334 & 0.8199 & \colorbox{green!20}{0.8457} & 0.8199 & \colorbox{red!20}{0.5605} \\
    \cdashlinelr{1-7}
    \multicolumn{7}{l}{\textbf{\methodname{}}} \\
    {Training system(s)} \\
    \Towervtwo{} Greedy & \colorbox{gray!20}{0.8278} & 0.8356 & 0.8199 & \colorbox{green!20}{0.8512} & 0.8225 & \colorbox{red!20}{0.6074} \\
    \claudethreefive{} & 0.8263 & 0.8338 & 0.8199 & \colorbox{green!20}{0.8615} & 0.8205 & \colorbox{red!20}{0.6014} \\
    All other models & \textbf{0.8347} & \textbf{0.8356} & \textbf{0.8199} & \colorbox{green!20}{\textit{\textbf{0.8656}}} & \textbf{0.8236} & \colorbox{red!20}{\textbf{0.6505}} \\
    \bottomrule
\end{tabular}
    \end{center}
    \caption{System-level soft pairwise accuracy for variations of \autorank{} and \methodname{} on the WMT24 for all system pairs containing \Towervtwo{}.
    On the columns, we vary the metrics used in each method's ensemble; on the rows, we vary the systems considered for training \methodname{}.
    The goal is to assess how sensitive the performance of methods are to changes in their ensembling / training procedures.
    Values in \colorbox{gray!20}{gray} serve as baselines, taken from Table~\ref{tab:metric-adjust-sys-ins}. Values in \colorbox{green!20}{green} and \colorbox{red!20}{red} represent the best and worst metric configurations, respectively.
    Values in \textbf{bold} are the best training system configuration for \methodname{}, for each metric configuration.
    The value in \textit{italic} is the best overall.
    }
    \label{tab:adjust-metric-variants}
\end{table*}

\paragraph{Other metrics do not necessarily outperform the interfering metric.}
Remarkably, the interfering metric, \comet{}, performs similarly to alternative metrics like \xcomet{} and \metricx{} on both sets of system pairs.
Thus, in this case, using other metrics in isolation for evaluation would not even achieve better results than evaluating with the interfering metric.
Moreover, \autorank{}, an ensembling baseline, does not seem to perform particularly well either.
Instead, our proposed instance-level variant of \autorank{}, \autorank{}-Ins, performs better, underscoring the importance of preserving instance-level information to maximize system-level performance.
%
% Apart from ensembling, preserving instance-level information is important to maximize system-level performance, which is consistent with the hypothesis we laid out in Section~\ref{sec:analysis-mbr-ii}.

\paragraph{\methodname{} and \autorank{} can be built from a reduced set of metrics.}
In Section~\ref{sec:analysis}, we reported that metrics react differently to \mint{} according to their kind.
For example, neural metrics usually agreed with neural interfering metrics, while lexical metrics did not.
These disagreements are valuable information that should not be discarded; they might help making more accurate predictions as to whether a translation \textit{really} improved in quality.
With this in mind, we leveraged a diverse set of metrics in our main results for learning \methodname{} and for computing \autorank{}.
However, it could be the case that certain sets of metrics work better.
We assess this in in Table~\ref{tab:adjust-metric-variants}, where we report the average system-level SPA of \methodname{} variants and \autorank{} built with distinct features (excluding \comet{}): 1) the same as Table~\ref{tab:metric-adjust-sys-ins}; 2) every metric in Table~\ref{tab:sys-level-filter}; 3) all QE metrics; 4) all reference-based metrics; 5) all neural metrics; 6) all lexical metrics.
Variants built from lexical metrics perform visibly worse, while the other variants are more similar to the baseline.
The signal from neural metrics seems important for learning \methodname{} on \comet{} (a neural metric itself), even though they are also potentially biased at inference time.\footnote{Lexical metrics may also be biased \textit{against} the interfering metric. Quantifying what part of the difference among metrics is \textit{real} and what is bias is interesting for future work.}
Notably, using only reference-based metrics surpasses the baseline considerably, showing that there is some room for optimization for both \autorank{} and \methodname{}.

\paragraph{\methodname{} can be learned from a diverse set of models.}
We learned \methodname{} from greedy-sampled translations of \Towervtwo{} because it was the most immediate option: we adjust the \comet{} MBR scores with the scores of the corresponding greedy-decoding system.
However, during evaluation we may not have access to translations with and without \mint{} for the same system; this was the case with \Towervtwo{} for the organizers of the shared task.
In the worst case, an unbiased equivalent to a biased system may not exist or be prohibitively hard to obtain (e.g., to a model pre-trained on filtered translation data).
Thus, we assessed whether \methodname{} could be learned from the translations of other systems in Table~\ref{tab:adjust-metric-variants}, which contains three versions of \methodname{}, each trained on translations from distinct systems: 1) \Towervtwo{} greedy; 2) \claudethreefive{}; 3) all systems available to the shared task organizers.
Results are similar across variants, with the variant learned from all models available performing the best.
This finding is encouraging for three reasons: 1) generalizability: \methodname{} can be strong even when learned from a single other system; 2) practicality: the non-\mint{} equivalent of the \mint{} system is not required to build \methodname{};  3) scalability: better results can be achieved by obtaining more translations from other systems.

\section{Related Work}\label{sec:related-work}
\subsection{Metric Interference.}
The role of automatic metrics has expanded beyond evaluation to applications in model training (e.g., as learning objectives in minimum error rate training)~\citep{och2003minimum,zaidan2009feasibility}, in reward modeling for human-preference alignment~\citep{Shu2021RewardOF,xucontrastive,zhu2024preference}, in training data filtering~\citep{alves2024tower,rei2024tower}, and in quality-aware decoding~\citep{fernandes2022quality,wu2024better}, where a metric is used to choose the best candidate out of a pool of model generations.
The aforementioned applications give rise to \textbf{Metric Interference (\mint{})}---formalized in Section~\ref{sec:metric-interference}---where such metrics are used for both optimization and evaluation. 
\mint{} introduces additional evaluation challenges due to inherent metric flaws and biases. 
For example, LLM judges tend to favor outputs from their own model family~\citep{panickssery2024llm,verga2024replacing}; in machine translation, this is also the case according to~\citet{agrawal-etal-2024-automatic-metrics}.
It has been long known that traditional metrics suffer from similar issues.
\citet{CallisonBurch2006ReevaluatingTR} show how optimizing a translation system directly for \bleu{} need not lead to improvements in translation quality according to humans.
This is because of inherent shortcomings of the metric, for example its insensitivity to synonyms.
Similarly, even models optimized for stronger metrics like \bleurt{} can produce absurd translations that score highly~\citep{yan2023bleurt}. 
Furthermore, \citet{fernandes2022quality} show that MBR and translation re-ranking with neural metrics lead to overestimated scores at a system level, reducing their correlation with human judgments and lexical metrics.
\citet{gisserot2024preference} report similar findings with Contrastive Preference Optimization~\citep{xucontrastive}, a method whereby models are trained to produce translations preferred by automatic metrics. 
Some of these works recommend using a different metric for evaluation than that used for optimization.
We find such strategies are insufficient to mitigate \mint{}, and propose an alternative.
%
%Our work analyzes the instance-level impact of \mint{} on metric scores, providing granular insights into the system-level dynamics the latter work reports, and proposes mitigation strategies for evaluation.

\subsection{Interference Mitigation Strategies.} 
For general-purpose evaluation of LLMs, \citet{verga2024replacing} propose using an ensemble of LLM judges to reduce intra-model bias (i.e., models preferring their own outputs).
For MT, \citet{kocmi-etal-2024-navigating} study how metric improvements relate to the probability of humans preferring the improved translation.
While they do not study the effect of \mint{}, they warn against its effects, and recommend not using interfering metrics for evaluation.
Other existing works that deal with \mint{} more directly attempt to improve the underlying optimization rather than the evaluation process.
\citet{kamigaito2024theoretical}
improve the reliability of MBR by averaging the scores yielded by the same metric trained from different initializations.
\citet{kovacsmitigating}~show that MBR metrics prefer their own systems or those of correlated metrics.
To address this, they perform MBR with an ensemble of metrics---as opposed to a single one---showing better correlation with human judgements.
Instead of trying to improve MBR, the goal of our proposed method is to recover more accurate metric scores to make evaluation more reliable under cases of metric interference.

\subsection{Broader Context on Metric Interference}
Goodhart's law states that ``when a
measure becomes a target, it ceases to be a good measure''.
The notion that relying too much on a limited set of evaluation methods can lead to the gaming thereof has been acknowledged in machine learning (ML) literature for long~\citep{Hutchinson2022EvaluationGI}.
And this over-reliance need not imply optimizing explicitly for automatic metrics, as in the use-cases our work focuses on. 
Simply using the same benchmarks and metrics repeatedly over time can deteriorate their value as proxies of human preference~\citep{Hutchinson2022EvaluationGI}.
Indeed, through comprehensive surveys that span various ML subfields (e.g., computer vision, NLP, recommender systems, RL, graph processing) \citet{Zhang2019MachineLT} and \cite{Liao2021AreWL} find a consistent trend of overfitting to a small amount of benchmarks, and discuss how this reduces the capability of automatic evaluation to measure scientific progress.
Furthermore, besides the aforementioned works on NLP and MT, similar patterns of metric gaming through overuse or explicit optimization have been found in applications of ranking~\citep{Yilmaz2010OnTC} and computer vision~\citep{Carter2020OverinterpretationRI,Cardelino2013ACS,Jiang2019TIGErTG,MaierHein2022MetricsRR}.
In most cases, the solution that is either explicitly proposed or hinted at is the same: more diverse and representative benchmarking methodologies---be they test sets or metrics---should be developed to ensure the continued relevance and credibility of automatic evaluation.

\section{Discussion on the Implications of Metric Interference}
\mint{} poses a threat to the integrity of evaluation frameworks; shared tasks and leaderboards where participants may tune models to optimize correlated metrics are prominent examples.
As discussed in Section~\ref{sec:related-work}, a common solution across various subfields of ML for this type of issue is to avoid relying on a small amount of evaluation methods.
As an ensemble method that can leverage an arbitrary combination of MT metrics, \methodname{} is aligned with this idea.

However, this may be a partial solution; other interventions could involve changing the design of evaluation protocols themselves, particularly in how optimization and evaluation metrics are selected and separated.
For example, shared task organizers could partition metrics into disjoint groups: some available during model development, others held out for evaluation.
Ideally, metrics chosen for optimization should be orthogonal to those used for evaluation, minimizing the risk of indirect overfitting. 
However, in practice, this separation is rarely clean, as many metrics are spuriously correlated: as shown in our analysis, optimization of one leads to measurable gains on others but not necessarily on human performance. 
This suggests that simply banning the use of evaluation metrics during training is insufficient unless their relationships can be robustly characterized and accounted for.
Another option could be to foster the design of automatic metrics that are ``aware'' they are being gamed.
An interesting first step in this direction could be to study how uncertainty relates to metric gaming, perhaps through analyzing how uncertainty-aware evaluation metrics~\citep{glushkova2021uncertainty} behave under metric interference.

These are open challenges, and further research requires the existence of more datasets where metric interference and human judgements are present.
This is the primary resource required to assess the effects of \mint{} on correlations of automatic metrics with human judgements.
In addressing metric interference explicitly, we hope this work encourages a broader rethinking of evaluation design towards setups that are robust to strategic optimization.

\section{Conclusion}
We analyzed two common cases of \mint{} (defined in Section~\ref{sec:metric-interference}) in MT---training data filtering and MBR---highlighting its negative effects on system- and instance-level evaluation with a myriad of metrics.
We find that interfering metric instance-level scores are distorted under \mint{}, with large improvements becoming more often associated to deterioration in the scores of other metrics and humans.
Importantly, such distortions hold for metrics that correlate with the interfering metric, bringing into question the common practice of using a single different metric for evaluation.
%
%Building on these insights, we propose a method for improving evaluation in cases of \mint{}: \methodname{}.
%
%At a system-level, our findings for MBR corroborate those of past literature, and we add insight on data filtering, which seems to be a less harsh case of \mint{}.
%
%Crucially, we provide novel analysis at an instance-level, providing insights on the inner workings of \mint{}.
%
%Relative to metric-agnostic improvements to models (e.g., scaling up model size), \mint{} causes metrics to disagree on the quality delta of many more instances.
%
%However, these remain minority as metrics still agree on most instances, with the quality of a significant portion even deteriorating according to the interfering metric.
%
%Importantly, these effects propagate to metrics that correlate with the interfering metric, which jeopardizes evaluation with those metrics.
%
%We believe this phenomenon drove the conflicting results between automatic and human evaluation found by \citet{kocmi2024findings}.

Based on these insights, we develop \methodname{}, a method for correcting the interfering metric during evaluation.
\methodname{} outperforms the interfering metric, other state-of-the art metrics, and the ensembling method used in the \wmttwofour{} MT shared task at ranking systems and translations. %particularly those from strong systems.

%In the future, we would like to explore the effects of \mint{} on the interpretation of metric values, such as the study by~\citet{kocmi-etal-2024-navigating}
In the future, it would be interesting to explore more cases of \mint{} (e.g., preference optimization) on more metrics and tasks.
%and its impact on the interpretation of metric scores.
%
That said, despite the importance of \mint{}, we note a lack of human evaluation data necessary to perform such studies and highlight the need to create such resources.
%
%We believe more research is necessary to fully understand \mint{} from theoretical and practical perspectives, and to devise reliable evaluation strategies for systems that leverage evaluation metrics.

\section*{Acknowledgements} 
We thank Sweta Agrawal, Amin Farajian, and Patrick Fernandes for their constructive feedback on the paper. We acknowledge EuroHPC JU for awarding the project ID EHPC-AI-2024A01-085 access to MareNostrum 5 ACC. This work was supported by EU’s Horizon Europe Research and Innovation Actions (UTTER, contract 101070631), by the project DECOLLAGE (ERC-2022-CoG 101088763), by the Portuguese Recovery and Resilience Plan through project C64500888200000055 (Center for Responsible AI), and by Fundação para a Ciência e Tecnologia through contract UIDB/50008/2020.

\bibliography{anthology_0,anthology_1,custom}
\bibliographystyle{tacl_natbib}
% \bibliography{custom}

\onecolumn
\appendix
\section{Metric details}\label{apx:metric-details}
We list below the metric implementations and checkpoints used in this paper:
\begin{itemize}

\item \textbf{\chrf{}}: \texttt{sacrebleu==2.4.2}, default settings. 
\item \textbf{\bleu{}}: \texttt{sacrebleu==2.4.2}, default settings.
\item \textbf{\comet{}}: Huggingface checkpoint: \texttt{Unbabel/wmt22-comet-da}.
\item \textbf{\cometkiwi{}}: Huggingface checkpoint: \texttt{Unbabel/wmt22-cometkiwi-da}.
\item \textbf{\xcomet{}}: Huggingface checkpoint: \texttt{Unbabel/XCOMET-XL}
\item \textbf{\metricx{}}: Huggingface checkpoint: \texttt{google/metricx-23-xl-v2p0}.
\item \textbf{\bleurt{}}: Huggingface checkpoint: \texttt{lucadiliello/BLEURT-20}.
\item \textbf{\prometheus{}}: Huggingface checkpoint: \texttt{prometheus-eval/prometheus-7b-v2.0}
\item \textbf{\iscore{}}: Huggingface checkpoint: \texttt{xu1998hz/InstructScore}.
\end{itemize}

\section{Random Forest Hyperparameters}\label{apx:rf-hp}
We use the default hyperparameters of the \href{https://scikit-learn.org/stable/modules/generated/sklearn.ensemble.RandomForestRegressor.html}{scikit-learn} implementation, except for the number of estimators and max depth. Namely: n\_estimators=100, *, criterion=``squared\_error'', max\_depth=None, min\_samples\_split=2, min\_samples\_leaf=1, min\_weight\_fraction\_leaf=0.0, max\_features=1.0, max\_leaf\_nodes=None, min\_impurity\_decrease=0.0, bootstrap=True, oob\_score=False, n\_jobs=None, random\_state=None, verbose=0, warm\_start=False, ccp\_alpha=0.0, max\_samples=None, monotonic\_cst=None

%\newpage

\section{Supplementary results}\label{apx:supp-df-mbr}

\begin{table}[h!]
    \centering
\footnotesize
\begin{tabular}{lcccc}
    \toprule
    Filtering & Random & CKiwi & xC & \textbf MX \\
    \midrule
    Random & - & - & - & - \\
    CKiwi & 0.0042 & - & - & - \\
    xC & 0.0040 & 0.0293 & - & - \\
    MX & 0.0042 & 0.0307 & 0.0002 & - \\
    \bottomrule
\end{tabular}
    \caption{Jaccard Index ($\frac{A \cap B}{A \cup B}$, where $A$ and $B$ are filtered training sets) among training instances selected by different metrics. 0 means no overlap.}
    \label{tab:filters-overlap}
\end{table}

%% DATA FILTERING
\begin{table*}[h]
    \begin{center}
    \setlength{\tabcolsep}{3pt}
\renewcommand{\arraystretch}{0.5}
\tiny
\begin{tabular}{l cc c@{\hspace{.4cm}} ccccccc}
    \toprule
    & \multicolumn{2}{c}{\textbf{Lexical}} & & \multicolumn{7}{c}{\textbf{Neural}} \\
    Filtering & \chrf{}$\uparrow$ & \bleu{}$\uparrow$ & & \comet{}$\uparrow$ & \bleurt{}$\uparrow$ & \cometkiwi{}$\uparrow$ & xC-REF$\uparrow$ & xC-QE$\uparrow$ & MX-REF$\downarrow$ & MX-QE$\downarrow$ \\
    \midrule
    Random & 50.75 & 21.03 & & 69.34 & 53.61 & 65.76 & 62.34 & 61.23 & 5.32 & 6.65 \\
    \cdashlinelr{1-11}
    \cometkiwi{} & \colorbox{green!20}{53.25} & \colorbox{green!20}{24.17} & & \colorbox{green!20}{73.61} & \colorbox{green!20}{57.83} & \colorbox{green!20}{\fbox{70.63}} & \colorbox{green!20}{70.47} & \colorbox{green!20}{69.72} & \colorbox{green!20}{4.11} & \colorbox{green!20}{4.90} \\
    \xcomet{} & \colorbox{red!20}{39.61} & \colorbox{red!20}{16.17} & & \colorbox{red!20}{65.46} & \colorbox{green!20}{53.97} & \colorbox{red!20}{59.18} & \colorbox{green!20}{64.91} & \colorbox{green!20}{\fbox{63.00}} & \colorbox{green!20}{5.29} & \colorbox{red!20}{10.06} \\
    \metricx{} & \colorbox{red!20}{48.39} & \colorbox{red!20}{19.32} & & \colorbox{green!20}{72.17} & \colorbox{green!20}{57.51} & \colorbox{green!20}{68.15} & \colorbox{green!20}{69.85} & \colorbox{green!20}{68.98} & \colorbox{green!20}{3.99} & \colorbox{green!20}{\fbox{5.22}} \\
    \bottomrule
\end{tabular}
    \end{center}
    \vspace{-1\baselineskip} 
    \caption{Data filtering results table like Table~\ref{tab:sys-level-filter} for English$\rightarrow$German.
    }
    \label{tab:apx-df-en-de}
\end{table*}
\vspace{-3em}

\begin{table*}[h]
    \begin{center}
    \setlength{\tabcolsep}{3pt}
\renewcommand{\arraystretch}{0.5}
\tiny
\begin{tabular}{l cc c@{\hspace{.4cm}} ccccccc}
    \toprule
    & \multicolumn{2}{c}{\textbf{Lexical}} & & \multicolumn{7}{c}{\textbf{Neural}} \\
    Filtering & \chrf{}$\uparrow$ & \bleu{}$\uparrow$ & & \comet{}$\uparrow$ & \bleurt{}$\uparrow$ & \cometkiwi{}$\uparrow$ & xC-REF$\uparrow$ & xC-QE$\uparrow$ & MX-REF$\downarrow$ & MX-QE$\downarrow$ \\
    \midrule
    Random & 55.69 & 28.86 & & 79.17 & 64.71 & 73.66 & 77.39 & 80.22 & 4.60 & 5.94 \\
    \cdashlinelr{1-11}
    \cometkiwi{} & \colorbox{green!20}{57.13} & \colorbox{green!20}{32.78} & & \colorbox{green!20}{80.10} & \colorbox{green!20}{65.46} & \colorbox{green!20}{\fbox{75.08}} & \colorbox{green!20}{80.90} & \colorbox{green!20}{83.24} & \colorbox{green!20}{4.49} & \colorbox{green!20}{5.77} \\
    \xcomet{} & \colorbox{red!20}{52.37} & \colorbox{red!20}{28.26} & & \colorbox{red!20}{76.93} & \colorbox{red!20}{62.53} & \colorbox{red!20}{72.44} & \colorbox{green!20}{78.08} & \colorbox{red!20}{\fbox{79.51}} & \colorbox{red!20}{5.05} & \colorbox{red!20}{7.71} \\
    \metricx{} & \colorbox{red!20}{55.50} & \colorbox{green!20}{31.36} & & \colorbox{red!20}{77.85} & \colorbox{red!20}{63.55} & \colorbox{red!20}{73.19} & \colorbox{red!20}{77.25} & \colorbox{red!20}{78.63} & \colorbox{red!20}{5.56} & \colorbox{red!20}{\fbox{7.34}} \\
    \bottomrule
\end{tabular}
    \end{center}
    \vspace{-1\baselineskip} 
    \caption{Data filtering results table like Table~\ref{tab:sys-level-filter} for German$\rightarrow$English.
    }
    \label{tab:apx-df-de-en}
\end{table*}
\vspace{-3em}

\begin{table*}[h]
    \begin{center}
    \setlength{\tabcolsep}{3pt}
\renewcommand{\arraystretch}{0.5}
\tiny
\begin{tabular}{l cc c@{\hspace{.4cm}} ccccccc}
    \toprule
    & \multicolumn{2}{c}{\textbf{Lexical}} & & \multicolumn{7}{c}{\textbf{Neural}} \\
    Filtering & \chrf{}$\uparrow$ & \bleu{}$\uparrow$ & & \comet{}$\uparrow$ & \bleurt{}$\uparrow$ & \cometkiwi{}$\uparrow$ & xC-REF$\uparrow$ & xC-QE$\uparrow$ & MX-REF$\downarrow$ & MX-QE$\downarrow$ \\
    \midrule
    Random & 38.67 & 13.09 & & 73.85 & 58.70 & 72.34 & 72.73 & 78.01 & 4.19 & 4.15 \\
    \cdashlinelr{1-11}
    \cometkiwi{} & \colorbox{green!20}{40.12} & \colorbox{green!20}{14.69} & & \colorbox{green!20}{76.30} & \colorbox{green!20}{61.11} & \colorbox{green!20}{\fbox{75.27}} & \colorbox{green!20}{78.47} & \colorbox{green!20}{84.02} & \colorbox{green!20}{3.23} & \colorbox{green!20}{2.87} \\
    \xcomet{} & \colorbox{red!20}{37.36} & \colorbox{green!20}{13.27} & & \colorbox{green!20}{74.05} & \colorbox{green!20}{58.96} & \colorbox{green!20}{72.39} & \colorbox{green!20}{76.97} & \colorbox{green!20}{\fbox{82.23}} & \colorbox{green!20}{3.58} & \colorbox{green!20}{3.43} \\
    \metricx{} & \colorbox{green!20}{38.90} & \colorbox{green!20}{14.58} & & \colorbox{green!20}{74.77} & \colorbox{green!20}{58.97} & \colorbox{green!20}{72.96} & \colorbox{green!20}{76.65} & \colorbox{green!20}{81.37} & \colorbox{green!20}{3.74} & \colorbox{green!20}{\fbox{3.41}} \\
    \bottomrule
\end{tabular}
    \end{center}
    \vspace{-1\baselineskip} 
    \caption{Data filtering results table like Table~\ref{tab:sys-level-filter} for Chinese$\rightarrow$English.
    }
    \label{tab:apx-df-zh-en}
\end{table*}
\vspace{-3em}

\begin{table*}[h]
    \begin{center}
    \setlength{\tabcolsep}{3pt}
\renewcommand{\arraystretch}{0.5}
\tiny
\begin{tabular}{l cc c@{\hspace{.4cm}} ccccccc}
    \toprule
    & \multicolumn{2}{c}{\textbf{Lexical}} & & \multicolumn{7}{c}{\textbf{Neural}} \\
    Filtering & \chrf{}$\uparrow$ & \bleu{}$\uparrow$ & & \comet{}$\uparrow$ & \bleurt{}$\uparrow$ & \cometkiwi{}$\uparrow$ & xC-REF$\uparrow$ & xC-QE$\uparrow$ & MX-REF$\downarrow$ & MX-QE$\downarrow$ \\
    \midrule
    Greedy & 68.19 & 42.61 & & 84.90 & 71.87 & 80.74 & 86.45 & 85.76 & 1.71 & 2.01 \\
    \cdashlinelr{1-11}
    \multicolumn{9}{l}{ \textbf{MBR}} \\
    $\chrf{}_{L}$ & \colorbox{green!20}{\fbox{68.22}} & \colorbox{red!20}{41.57} & & \colorbox{green!20}{84.92} & \colorbox{red!20}{71.85} & \colorbox{green!20}{80.78} & \colorbox{red!20}{86.15} & \colorbox{red!20}{85.36} & \colorbox{green!20}{1.67} & \colorbox{green!20}{1.97} \\
    $\bleu{}_{L}$ & \colorbox{red!20}{67.53} & \colorbox{red!20}{\fbox{42.07}} & & \colorbox{red!20}{84.68} & \colorbox{red!20}{71.45} & \colorbox{red!20}{80.58} & \colorbox{red!20}{86.23} & \colorbox{red!20}{85.37} & \colorbox{red!20}{1.76} & \colorbox{red!20}{2.07} \\
    \cdashlinelr{1-11}
    $\comet{}_{N}$ & \colorbox{red!20}{66.94} & \colorbox{red!20}{39.99} & & \colorbox{green!20}{\fbox{86.21}} & \colorbox{green!20}{72.86} & \colorbox{green!20}{81.45} & \colorbox{green!20}{87.52} & \colorbox{green!20}{86.92} & \colorbox{green!20}{1.49} & \colorbox{green!20}{1.71} \\
    $\bleurt{}_{N}$ & \colorbox{red!20}{66.44} & \colorbox{red!20}{39.29} & & \colorbox{green!20}{85.20} & \colorbox{green!20}{\fbox{74.26}} & \colorbox{green!20}{81.24} & \colorbox{green!20}{87.09} & \colorbox{green!20}{86.57} & \colorbox{green!20}{1.44} & \colorbox{green!20}{1.65} \\
    \bottomrule
\end{tabular}
    \end{center}
    \vspace{-1\baselineskip} 
    \caption{MBR results table like Table~\ref{tab:sys-level-mbr} English$\rightarrow$German.
    }
    \label{tab:apx-mbr-en-de}
\end{table*}
\vspace{-2em}

\begin{table*}[h]
    \begin{center}
    \setlength{\tabcolsep}{3pt}
\renewcommand{\arraystretch}{0.5}
\tiny
\begin{tabular}{l cc c@{\hspace{.4cm}} ccccccc}
    \toprule
    & \multicolumn{2}{c}{\textbf{Lexical}} & & \multicolumn{7}{c}{\textbf{Neural}} \\
    Filtering & \chrf{}$\uparrow$ & \bleu{}$\uparrow$ & & \comet{}$\uparrow$ & \bleurt{}$\uparrow$ & \cometkiwi{}$\uparrow$ & xC-REF$\uparrow$ & xC-QE$\uparrow$ & MX-REF$\downarrow$ & MX-QE$\downarrow$ \\
    \midrule
    Greedy & 69.65 & 47.20 & & 86.02 & 75.26 & 80.03 & 89.94 & 91.07 & 2.34 & 2.37 \\
    \cdashlinelr{1-11}
    \multicolumn{9}{l}{ \textbf{MBR}} \\
    $\chrf{}_{L}$ & \colorbox{green!20}{\fbox{69.67}} & \colorbox{red!20}{46.32} & & \colorbox{red!20}{86.00} & \colorbox{red!20}{75.13} & \colorbox{green!20}{80.07} & \colorbox{red!20}{89.90} & \colorbox{red!20}{91.07} & \colorbox{green!20}{2.33} & \colorbox{green!20}{2.32} \\
    $\bleu{}_{L}$ & \colorbox{red!20}{69.63} & \colorbox{green!20}{\fbox{47.54}} & & \colorbox{green!20}{86.07} & \colorbox{red!20}{75.26} & \colorbox{green!20}{80.04} & \colorbox{green!20}{90.02} & \colorbox{red!20}{91.06} & \colorbox{red!20}{2.35} & \colorbox{green!20}{2.36} \\
    \cdashlinelr{1-11}
    $\comet{}_{N}$ & \colorbox{red!20}{68.50} & \colorbox{red!20}{45.08} & & \colorbox{green!20}{\fbox{86.76}} & \colorbox{green!20}{75.51} & \colorbox{green!20}{80.24} & \colorbox{green!20}{90.47} & \colorbox{green!20}{91.40} & \colorbox{green!20}{2.17} & \colorbox{green!20}{2.19} \\
    $\bleurt{}_{N}$ & \colorbox{red!20}{69.05} & \colorbox{red!20}{45.63} & & \colorbox{green!20}{86.34} & \colorbox{green!20}{\fbox{76.18}} & \colorbox{green!20}{80.20} & \colorbox{green!20}{90.32} & \colorbox{green!20}{91.30} & \colorbox{green!20}{2.17} & \colorbox{green!20}{2.21} \\
    \bottomrule
\end{tabular}
    \end{center}
    \vspace{-1\baselineskip} 
    \caption{MBR results table like Table~\ref{tab:sys-level-mbr} German$\rightarrow$English.
    }
    \label{tab:apx-mbr-de-en}
\end{table*}
\vspace{-2em}

\begin{table*}[h]
    \begin{center}
    \setlength{\tabcolsep}{3pt}
\renewcommand{\arraystretch}{0.5}
\tiny
\begin{tabular}{l cc c@{\hspace{.4cm}} ccccccc}
    \toprule
    & \multicolumn{2}{c}{\textbf{Lexical}} & & \multicolumn{7}{c}{\textbf{Neural}} \\
    Filtering & \chrf{}$\uparrow$ & \bleu{}$\uparrow$ & & \comet{}$\uparrow$ & \bleurt{}$\uparrow$ & \cometkiwi{}$\uparrow$ & xC-REF$\uparrow$ & xC-QE$\uparrow$ & MX-REF$\downarrow$ & MX-QE$\downarrow$ \\
    \midrule
    Greedy & 49.80 & 23.84 & & 81.58 & 69.15 & 80.79 & 87.51 & 90.18 & 2.17 & 2.00 \\
    \cdashlinelr{1-11}
    \multicolumn{9}{l}{ \textbf{MBR}} \\
    $\chrf{}_{L}$ & \colorbox{green!20}{\fbox{50.05}} & \colorbox{red!20}{23.25} & & \colorbox{green!20}{81.63} & \colorbox{green!20}{69.17} & \colorbox{green!20}{80.85} & \colorbox{red!20}{87.17} & \colorbox{red!20}{89.98} & \colorbox{red!20}{2.18} & \colorbox{red!20}{2.03} \\
    $\bleu{}_{L}$ & \colorbox{red!20}{49.29} & \colorbox{red!20}{\fbox{23.61}} & & \colorbox{red!20}{81.58} & \colorbox{red!20}{68.98} & \colorbox{red!20}{80.71} & \colorbox{red!20}{87.35} & \colorbox{red!20}{90.13} & \colorbox{red!20}{2.20} & \colorbox{red!20}{2.03} \\
    \cdashlinelr{1-11}
    $\comet{}_{N}$ & \colorbox{red!20}{48.87} & \colorbox{red!20}{22.71} & & \colorbox{green!20}{\fbox{82.32}} & \colorbox{green!20}{69.30} & \colorbox{green!20}{81.15} & \colorbox{green!20}{87.82} & \colorbox{green!20}{90.71} & \colorbox{green!20}{2.06} & \colorbox{green!20}{1.87} \\
    $\bleurt{}_{N}$ & \colorbox{red!20}{48.89} & \colorbox{red!20}{22.21} & & \colorbox{green!20}{81.71} & \colorbox{green!20}{\fbox{70.12}} & \colorbox{green!20}{81.07} & \colorbox{green!20}{87.97} & \colorbox{green!20}{90.59} & \colorbox{green!20}{2.03} & \colorbox{green!20}{1.85} \\
    \bottomrule
\end{tabular}
    \end{center}
    \vspace{-1\baselineskip} 
    \caption{MBR results table like Table~\ref{tab:sys-level-mbr} Chinese$\rightarrow$English.
    }
    \label{tab:apx-mbr-zh-en}
\end{table*}
\vspace{-2em}

\begin{table*}[h]
    \begin{center}
    \setlength{\tabcolsep}{3pt}
\renewcommand{\arraystretch}{0.5}
\tiny
\begin{tabular}{l cc c@{\hspace{.4cm}} ccccccc}
    \toprule
    & \multicolumn{2}{c}{\textbf{Lexical}} & & \multicolumn{7}{c}{\textbf{Neural}} \\
    Filtering & \chrf{}$\uparrow$ & \bleu{}$\uparrow$ & & \comet{}$\uparrow$ & \bleurt{}$\uparrow$ & \cometkiwi{}$\uparrow$ & xC-REF$\uparrow$ & xC-QE$\uparrow$ & MX-REF$\downarrow$ & MX-QE$\downarrow$ \\
    \midrule
    Greedy & 43.23 & 44.46 & & 87.22 & 72.89 & 81.23 & 87.27 & 84.31 & 1.40 & 1.34 \\
    \cdashlinelr{1-11}
    \multicolumn{9}{l}{ \textbf{MBR}} \\
    $\chrf{}_{L}$ & \colorbox{green!20}{\fbox{44.23}} & \colorbox{green!20}{44.91} & & \colorbox{green!20}{87.42} & \colorbox{green!20}{73.09} & \colorbox{green!20}{81.33} & \colorbox{red!20}{87.22} & \colorbox{red!20}{84.29} & \colorbox{green!20}{1.39} & \colorbox{green!20}{1.33} \\
    $\bleu{}_{L}$ & \colorbox{green!20}{44.02} & \colorbox{green!20}{\fbox{45.49}} & & \colorbox{green!20}{87.43} & \colorbox{green!20}{73.03} & \colorbox{red!20}{81.18} & \colorbox{red!20}{87.18} & \colorbox{red!20}{84.23} & \colorbox{red!20}{1.40} & \colorbox{red!20}{1.36} \\
    \cdashlinelr{1-11}
       $\comet{}_{N}$ & \colorbox{red!20}{40.35} & \colorbox{red!20}{40.93} & & \colorbox{green!20}{\fbox{88.72}} & \colorbox{red!20}{72.88} & \colorbox{green!20}{82.19} & \colorbox{green!20}{88.36} & \colorbox{green!20}{85.96} & \colorbox{green!20}{1.26} & \colorbox{green!20}{1.15} \\
    $\bleurt{}_{N}$ & \colorbox{red!20}{39.25} & \colorbox{red!20}{39.62} & & \colorbox{green!20}{87.28} & \colorbox{green!20}{\fbox{74.93}} & \colorbox{green!20}{81.80} & \colorbox{green!20}{88.10} & \colorbox{green!20}{85.64} & \colorbox{green!20}{1.25} & \colorbox{green!20}{1.15} \\
    \bottomrule
\end{tabular}
    \end{center}
    \vspace{-1\baselineskip} 
    \caption{MBR results table like Table~\ref{tab:sys-level-mbr} for 50 candidates.
    }
    \label{tab:apx-mbr-50-cands}
\end{table*}
\vspace{-2em}

\begin{table*}[h]
    \begin{center}
    \setlength{\tabcolsep}{3pt}
\renewcommand{\arraystretch}{0.5}
\tiny
\begin{tabular}{l cc c@{\hspace{.4cm}} ccccccc}
    \toprule
    & \multicolumn{2}{c}{\textbf{Lexical}} & & \multicolumn{7}{c}{\textbf{Neural}} \\
    Filtering & \chrf{}$\uparrow$ & \bleu{}$\uparrow$ & & \comet{}$\uparrow$ & \bleurt{}$\uparrow$ & \cometkiwi{}$\uparrow$ & xC-REF$\uparrow$ & xC-QE$\uparrow$ & MX-REF$\downarrow$ & MX-QE$\downarrow$ \\
    \midrule
    Greedy & 43.66 & 45.03 & & 87.32 & 72.90 & 81.22 & 87.73 & 84.59 & 1.36 & 1.33 \\
    \cdashlinelr{1-11}
    \multicolumn{9}{l}{ \textbf{MBR}} \\
    $\chrf{}_{L}$ & \colorbox{green!20}{\fbox{43.83}} & \colorbox{red!20}{44.40} & & \colorbox{green!20}{87.33} & \colorbox{red!20}{72.87} & \colorbox{green!20}{81.23} & \colorbox{red!20}{87.06} & \colorbox{red!20}{84.18} & \colorbox{red!20}{1.37} & \colorbox{green!20}{1.32} \\
    $\bleu{}_{L}$ & \colorbox{red!20}{43.17} & \colorbox{red!20}{\fbox{44.88}} & & \colorbox{red!20}{87.21} & \colorbox{red!20}{72.65} & \colorbox{red!20}{81.04} & \colorbox{red!20}{87.04} & \colorbox{red!20}{84.13} & \colorbox{red!20}{1.40} & \colorbox{red!20}{1.37} \\
    \cdashlinelr{1-11}
    $\comet{}_{N}$ & \colorbox{red!20}{40.05} & \colorbox{red!20}{40.71} & & \colorbox{green!20}{\fbox{88.53}} & \colorbox{green!20}{72.91} & \colorbox{green!20}{82.05} & \colorbox{green!20}{88.38} & \colorbox{green!20}{86.05} & \colorbox{green!20}{1.26} & \colorbox{green!20}{1.17} \\
    $\bleurt{}_{N}$ & \colorbox{red!20}{39.06} & \colorbox{red!20}{39.63} & & \colorbox{red!20}{87.21} & \colorbox{green!20}{\fbox{74.59}} & \colorbox{green!20}{81.66} & \colorbox{green!20}{88.09} & \colorbox{green!20}{85.77} & \colorbox{green!20}{1.24} & \colorbox{green!20}{1.18} \\
    \bottomrule
\end{tabular}
    \end{center}
    \vspace{-1\baselineskip} 
    \caption{MBR results table like Table~\ref{tab:sys-level-mbr} for \Towervtwo{}-70B.
    }
    \label{tab:apx-mbr-20-70b}
\end{table*}
\vspace{-2em}

\begin{table*}[h]
    \begin{center}
    \setlength{\tabcolsep}{4pt}
\renewcommand{\arraystretch}{1.2}
\tiny
\begin{tabular}{l ccccccccc}
    %\toprule
    \textbf{Metric} & C22 & CK22 & xCR & xCQ & MxR & MxQ & BT & \chrf{} & \bleu{} \\
    %\midrule
    \comet{} (C22) & \cellcolor{gray!25} & \cellcolor{red!28}0.560 & \cellcolor{red!31}0.623 & \cellcolor{red!20}0.409 & \cellcolor{red!35}0.692 & \cellcolor{red!24}0.470 & \cellcolor{red!38}0.752 & \cellcolor{red!34}0.673 & \cellcolor{red!33}0.665 \\
    \cometkiwi{} (CK22) & \cellcolor{red!28}0.560 & \cellcolor{gray!25} & \cellcolor{red!27}0.539 & \cellcolor{red!25}0.496 & \cellcolor{red!26}0.520 & \cellcolor{red!31}0.616 & \cellcolor{red!24}0.471 & \cellcolor{red!18}0.350 & \cellcolor{red!17}0.338 \\
    \xcomet{}-REF (xCR) & \cellcolor{red!31}0.623 & \cellcolor{red!27}0.539 & \cellcolor{gray!25} & \cellcolor{red!33}0.651 & \cellcolor{red!30}0.605 & \cellcolor{red!25}0.496 & \cellcolor{red!29}0.586 & \cellcolor{red!22}0.445 & \cellcolor{red!22}0.435 \\
    \xcomet{}-QE (xCQ) & \cellcolor{red!20}0.409 & \cellcolor{red!25}0.496 & \cellcolor{red!33}0.651 & \cellcolor{gray!25} & \cellcolor{red!21}0.411 & \cellcolor{red!23}0.468 & \cellcolor{red!18}0.359 & \cellcolor{red!12}0.242 & \cellcolor{red!11}0.228 \\
    \metricx{}-REF (MxR) & \cellcolor{red!35}0.692 & \cellcolor{red!26}0.520 & \cellcolor{red!30}0.605 & \cellcolor{red!21}0.411 & \cellcolor{gray!25} & \cellcolor{red!27}0.548 & \cellcolor{red!34}0.678 & \cellcolor{red!26}0.513 & \cellcolor{red!25}0.497 \\
    \metricx{}-QE (MxQ) & \cellcolor{red!24}0.470 & \cellcolor{red!31}0.616 & \cellcolor{red!25}0.496 & \cellcolor{red!23}0.468 & \cellcolor{red!27}0.548 & \cellcolor{gray!25} & \cellcolor{red!21}0.418 & \cellcolor{red!14}0.285 & \cellcolor{red!13}0.268 \\
    \bleurt{} (BT) & \cellcolor{red!38}0.752 & \cellcolor{red!24}0.471 & \cellcolor{red!29}0.586 & \cellcolor{red!18}0.359 & \cellcolor{red!34}0.678 & \cellcolor{red!21}0.418 & \cellcolor{gray!25} & \cellcolor{red!33}0.663 & \cellcolor{red!33}0.651 \\
    \chrf{} & \cellcolor{red!34}0.673 & \cellcolor{red!18}0.350 & \cellcolor{red!22}0.445 & \cellcolor{red!12}0.242 & \cellcolor{red!26}0.513 & \cellcolor{red!14}0.285 & \cellcolor{red!33}0.663 & \cellcolor{gray!25} & \cellcolor{red!47}0.937 \\
    \bleu{} & \cellcolor{red!33}0.665 & \cellcolor{red!17}0.338 & \cellcolor{red!22}0.435 & \cellcolor{red!11}0.228 & \cellcolor{red!25}0.497 & \cellcolor{red!13}0.268 & \cellcolor{red!33}0.651 & \cellcolor{red!47}0.937 & \cellcolor{gray!25} \\
    %\bottomrule
\end{tabular}
    \end{center}
    \vspace{-1\baselineskip} 
    \caption{Spearman correlation between evaluation metrics on the \wmttwothree{} English$\rightarrow$Chinese test set. Values are computed by averaging the correlations among scores for every unique source.}
    \label{tab:apx-corr-matrix}
\end{table*}
\vspace{-2em}

%\section{\methodname{} results broken down by LP}\label{apx:lp-results-metricadjust}

%\begin{table*}[h]
%    \begin{center}
%    \include{tables/appendix_mintadjust/all_sys_lp}
%    \end{center}
%    \vspace{-1\baselineskip} 
%    \caption{System-level SPA results for all LPs (direction omitted for from-English LPs).}
%    \label{tab:apx-metric-adjust-lp-sys-all}
%\end{table*}
%\vspace{-2em}

\begin{table*}[h]
    \begin{center}
    \setlength{\tabcolsep}{3pt}
\renewcommand{\arraystretch}{0.5}
\tiny
\begin{tabular}{l ccccccccccc}
    \toprule
    Metrics & de & es & cs & ru & uk & is & ja & zh & hi & cs$\rightarrow{}$uk & ja$\rightarrow{}$zh \\
    \midrule
    \multicolumn{9}{l}{\scriptsize \bf Baselines} \\
    $\chrf{}_{L,REF}$ & 0.3443 & 0.4163 & 0.0193 & 0.0357 & 0.2720 & \textbf{0.9985} & \textbf{0.6993} & 0.2883 & 0.4050 & \textbf{0.7440} & \textbf{0.7883} \\
    $\cometkiwi{}_{N,QE}$ & 0.7543 & 0.5837 & 0.9960 & 0.9847 & 0.7317 & 0.5015 & 0.3013 & 0.7903 & 0.9243 & 0.3443 & 0.2117 \\
    $\xcomet{}_{N,REF}$ & 0.7543 & 0.5837 & 0.9960 & 0.9847 & 0.7317 & 0.5015 & 0.3013 & 0.7903 & 0.9243 & 0.3443 & 0.2117 \\
    $\metricx{}_{N,REF}$ & 0.7543 & 0.5837 & 0.9960 & 0.9847 & 0.7317 & 0.5015 & 0.3013 & 0.7903 & 0.9420 & 0.3773 & 0.2117 \\
    $\comet{}_{N,REF}*$ & 0.7543 & 0.5837 & 0.9960 & 0.9847 & 0.7317 & 0.5015 & 0.3013 & 0.7903 & 0.9243 & 0.6663 & 0.2183 \\
    $\autorank{}_{E}$ & 0.7543 & 0.5837 & 0.9960 & 0.9847 & 0.7317 & 0.5015 & 0.3013 & 0.7903 & 0.9243 & 0.3443 & 0.2117 \\
    $\metametric{}_{E}$ & 0.7543 & 0.5837 & 0.9960 & 0.9847 & 0.7317 & 0.5015 & 0.3013 &\textbf{ 0.8100} & 0.9247 & 0.3443 & 0.2247 \\
    \cdashlinelr{1-12}
    \multicolumn{9}{l}{\scriptsize \bf This work} \\
    \autorank{}-Ins$_{E}$ & 0.7543 & 0.5840 & 0.9960 & 0.9847 & 0.7317 & 0.5015 & 0.3047 & 0.7903 & \textbf{0.9250} & 0.6453 & 0.2153 \\
    $\methodname{}_{E}$ & \textbf{0.7577} & \textbf{0.5847} & \textbf{0.9960} & \textbf{0.9847} & \textbf{0.7317} & 0.5015 & 0.3160 & 0.7903 & 0.9243 & 0.6773 & 0.2660 \\
    \bottomrule
\end{tabular}
    \end{center}
    \vspace{-1\baselineskip} 
    \caption{System-level SPA results (high-quality systems) for all LPs (direction omitted for from-English LPs).}
    \label{tab:apx-metric-adjust-lp-sys-high-q}
\end{table*}
\vspace{-2em}

\begin{table*}[h]
    \begin{center}
    \setlength{\tabcolsep}{3pt}
\renewcommand{\arraystretch}{0.5}
\tiny
\begin{tabular}{l ccccccccccc}
    \toprule
    Metrics & de & es & cs & ru & uk & is & ja & zh & hi & cs$\rightarrow{}$uk & ja$\rightarrow{}$zh \\
    \midrule
    \multicolumn{9}{l}{\scriptsize \bf Baselines} \\
    $\chrf{}_{L,REF}$ & 0.3653 & 0.4704 & 0.4845 & 0.4466 & 0.3543 & 0.6071 & 0.5117 & 0.4860 & 0.4973 & 0.4286 & 0.3505 \\
    $\cometkiwi{}_{N,QE}$ & 0.3771 & 0.4642 & 0.5338 & 0.4748 & 0.4016 & 0.5752 & 0.4951 & \textbf{0.5634} & 0.5254 & 0.4319 & 0.3449 \\
    $\xcomet{}_{N,REF}$ & \textbf{0.4386} & 0.4839 & 0.5667 & 0.5053 & \textbf{0.4282} & 0.6479 & 0.5133 & 0.5455 & 0.5250 & 0.4374 & 0.3731 \\
    $\metricx{}_{N,REF}$ & 0.4322 & 0.4931 & 0.5340 & 0.5010 & 0.4241 & 0.6394 & 0.5041 & 0.5339 & 0.4984 & 0.4172 & 0.3774 \\
    $\comet{}_{N,REF}*$ & 0.4016 & 0.4916 & \textbf{0.5778} & \textbf{0.5129} & 0.4027 & 0.6668 & 0.5130 & 0.5365 & 0.5194 & 0.4501 & \textbf{0.3780} \\
    $\metametric{}_{E}$ & 0.3504 & 0.4484 & 0.5122 & 0.4463 & 0.3954 & 0.5908 & 0.4209 & 0.4660 & 0.4512 & 0.4252 & 0.3709 \\
    \cdashlinelr{1-12}
    \multicolumn{12}{l}{\scriptsize \bf This work} \\
    \autorank{}-Ins$_{E}$ & 0.3983 & 0.5030 & 0.5697 & 0.5059 & 0.4009 & \textbf{0.6742} & \textbf{0.5195} & 0.5481 & \textbf{0.5294} & \textbf{0.4523} & 0.3745 \\
    $\methodname{}_{E}$ & 0.4172 & \textbf{0.5038} & 0.5683 & 0.5039 & 0.4049 & 0.6565 & 0.5095 & 0.5400 & 0.5223 & 0.4514 & 0.3740 \\
    \bottomrule
\end{tabular}
    \end{center}
    \vspace{-1\baselineskip} 
    \caption{Instance-level PA results for all LPs (direction omitted for from-English LPs).}
    \label{tab:apx-metric-adjust-lp-seg-all}
\end{table*}
\vspace{-2em}

\begin{table*}[h]
    \begin{center}
    \setlength{\tabcolsep}{3pt}
\renewcommand{\arraystretch}{0.5}
\tiny
\begin{tabular}{l ccccccccccc}
    \toprule
    Metrics & de & es & cs & ru & uk & is & ja & zh & hi & cs$\rightarrow{}$uk & ja$\rightarrow{}$zh \\
    \midrule
    \multicolumn{9}{l}{\scriptsize \bf Baselines} \\
    $\chrf{}_{L,REF}$ & 0.3182 & 0.4666 & 0.4608 & 0.4450 & 0.3454 & 0.5561 & \textbf{0.5219} & 0.4824 & 0.4781 & 0.4058 & 0.2806 \\
    $\cometkiwi{}_{N,QE}$ & 0.3128 & 0.4255 & 0.4813 & 0.4408 & 0.3691 & 0.5577 & 0.4781 & \textbf{0.5608} & 0.4950 & 0.4082 & 0.2506 \\
    $\xcomet{}_{N,REF}$ & \textbf{0.3759} & 0.4387 & 0.5064 & 0.4797 & 0.3913 & 0.5774 & 0.4924 & 0.5155 & 0.4971 & 0.4208 & 0.2656 \\
    $\metricx{}_{N,REF}$ & 0.3560 & 0.4602 & 0.4737 & 0.4860 & \textbf{0.4034} & 0.5600 & 0.4771 & 0.4924 & 0.4655 & 0.3975 & 0.2518 \\
    $\comet{}_{N,REF}*$ & 0.3347 & 0.4518 & \textbf{0.5193} & \textbf{0.4866} & 0.3802 & 0.5798 & 0.4760 & 0.4934 & 0.4860 & 0.4249 & 0.2710 \\
    $\metametric{}_{E}$ & 0.3299 & 0.4234 & 0.4760 & 0.4313 & 0.3749 & 0.5047 & 0.4055 & 0.4471 & 0.4271 & 0.4075 & \textbf{0.2992} \\
    \cdashlinelr{1-12}
    \multicolumn{12}{l}{\scriptsize \bf This work} \\
    \autorank{}-Ins$_{E}$ & 0.3237 & 0.4608 & 0.5105 & 0.4824 & 0.3760 & \textbf{0.5916} & 0.5034 & 0.5182 & 0.4961 & \textbf{0.4268} & 0.2710 \\
    $\methodname{}_{E}$ & 0.3491 & \textbf{0.4734} & 0.5164 & 0.4750 & 0.3865 & 0.5695 & \textbf{0.5008} & 0.5271 & \textbf{0.5008} & 0.4265 & 0.2812 \\
    \bottomrule
\end{tabular}
    \end{center}
    \vspace{-1\baselineskip} 
    \caption{Instance-level PA results (high-quality systems) for all LPs (direction omitted for from-English LPs).}
    \label{tab:apx-metric-adjust-lp-seg-high-q}
\end{table*}
\vspace{-2em}
\end{document}